\documentclass{article} 
\usepackage{iclr2027_conference,times}

\usepackage{amsmath,amsfonts,bm}

\def\eqref#1{equation~\ref{#1}}

\def\1{\bm{1}}

\DeclareMathAlphabet{\mathsfit}{\encodingdefault}{\sfdefault}{m}{sl}
\SetMathAlphabet{\mathsfit}{bold}{\encodingdefault}{\sfdefault}{bx}{n}

\usepackage[hidelinks]{hyperref}
\usepackage{url}
\usepackage{algorithm}
\usepackage{algorithmic}
\usepackage{booktabs}
\usepackage{tcolorbox}
\usepackage{listings}
\tcbuselibrary{breakable}
\usepackage{wrapfig}
\usepackage[table]{xcolor}
\usepackage{enumitem}
\usepackage{capt-of}

\usepackage{soul}
\sethlcolor{yellow!30}
\newcommand{\repchange}[1]{\hl{#1}}
\newcommand{\huggingface}{\includegraphics[height=0.3cm]{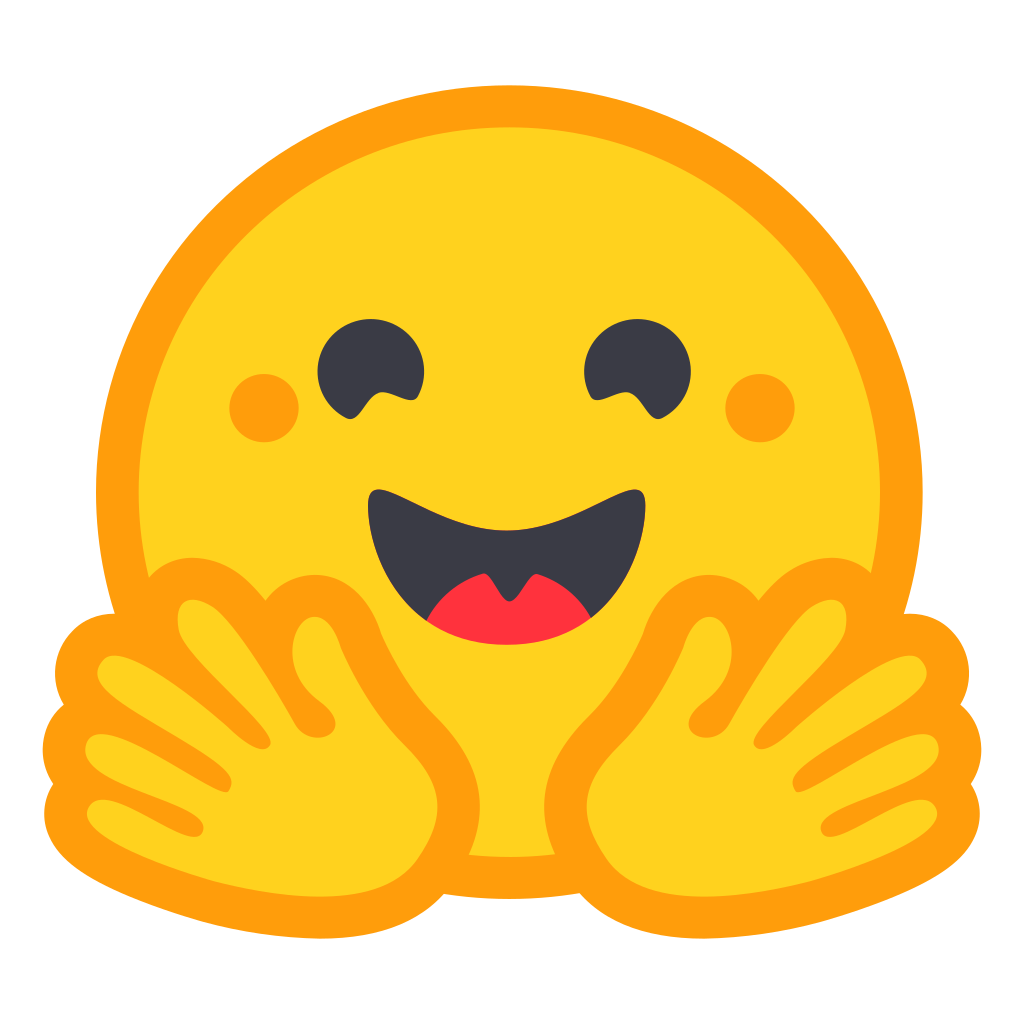}}

\title{Learning User Representations to Improve Simulation Fidelity}

\author{\textbf{Shuhaib Mehri\thanks{Work done while interning at Microsoft Research. Correspondence to: \href{mailto:mehri2@illinois.edu}{mehri2@illinois.edu}}~~$^{\spadesuit}$}$^{\diamondsuit}$,
    \textbf{Philippe Laban$^{\diamondsuit}$},
    \textbf{Michel Galley$^{\diamondsuit}$},
    \textbf{Dilek Hakkani-T\"ur$^{\spadesuit}$}, \\
    \textbf{Jennifer Neville$^{\diamondsuit}$},
    \textbf{Jianfeng Gao$^{\diamondsuit}$} \\
    \textnormal{$^{\spadesuit}$ University of Illinois Urbana-Champaign, $^{\diamondsuit}$} Microsoft Research \\
}

\iclrfinalcopy 
\begin{document}

\maketitle

\vspace{-1em}

\begin{abstract}
User simulators enable interactive training and evaluation of AI assistants at scale. Yet current simulators struggle to reproduce the diverse behaviors of real users, leading to assistants that generalize poorly to underrepresented users. We identify user representations as an underexplored bottleneck in user simulation. User representations are natural-language descriptions that condition simulators to reproduce a particular user's behavior. A representation's quality becomes apparent in hindsight, through the behavior it induces in a simulator. To evaluate and optimize user representations, we formulate a conversation reconstruction task: given an original conversation, we generate a representation, condition a simulator on it, simulate a conversation, and then measure fidelity by comparing user behavior in the original and simulated conversations. This comparison also yields \emph{hindsight feedback}, which identifies where the representation falls short and how it can be improved. We introduce \emph{hindsight distillation}, an online training procedure that distills feedback-conditioned improvements into the representation generator, enabling it to generate improved representations directly from the original conversation. We present \texttt{UserRepGen-4B}, a representation generator trained with hindsight distillation. Across 700 conversations spanning seven datasets, hindsight feedback helps improve user representations at test time, and \texttt{UserRepGen-4B} generates representations that enable more faithful simulation than those generated by GPT-5.5. Beyond conversation reconstruction, learned representations improve simulation fidelity for unseen tasks, better preserve the population-level diversity of real users, and better distinguish users across conversations.
\end{abstract}

\vspace{-0.5em}
\hspace{4.5cm} \href{https://huggingface.co/microsoft/UserRepGen-4B}{\huggingface~\texttt{microsoft/UserRepGen-4B}}

\begin{figure*}[h]
    \centering
    \includegraphics[width=\textwidth]{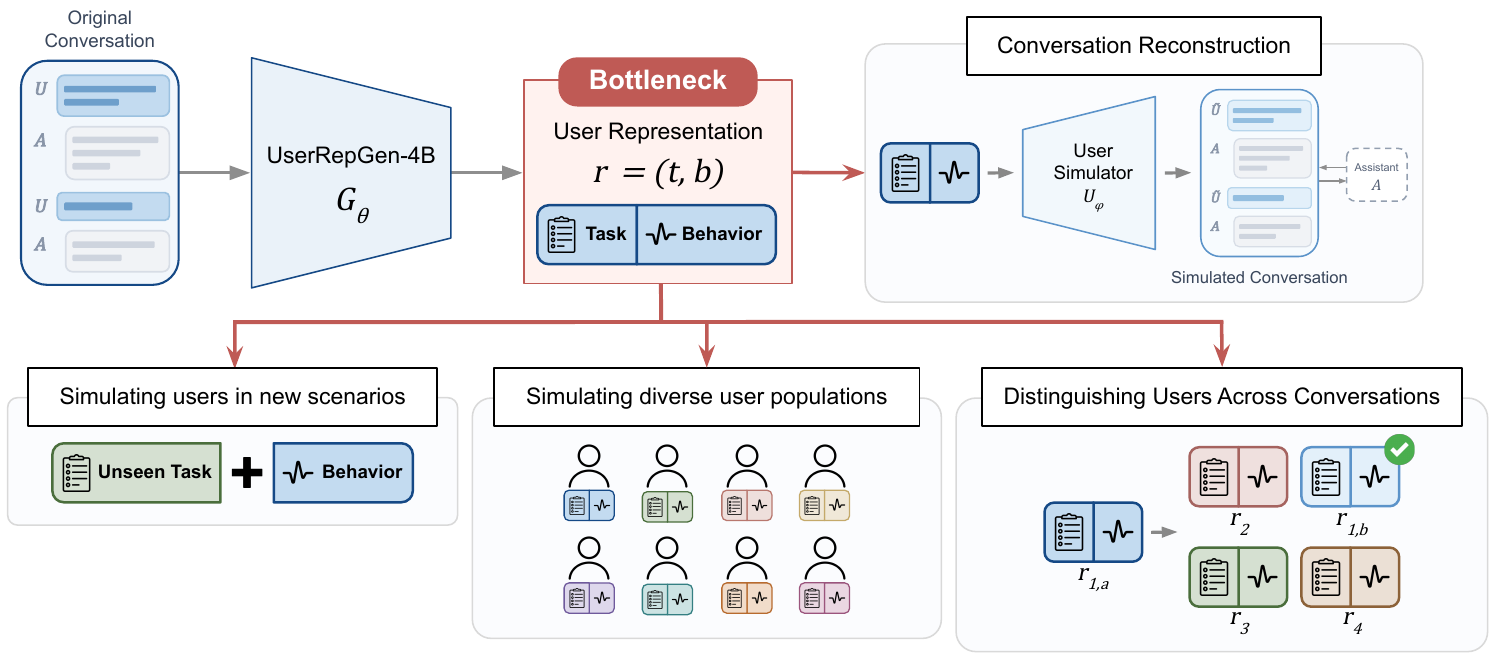}
    \vspace{-2em}
    \caption{\textbf{User representations are a bottleneck in user simulation.} They capture the information a simulator needs to faithfully reproduce a user's behavior. We train \texttt{UserRepGen-4B} to generate high-quality user representations and demonstrate their importance for reconstructing user conversations, simulating users in new scenarios, simulating diverse user populations, and distinguishing users across conversations.}
    \label{fig:intro}
    \vspace{-1em}
\end{figure*}

\section{Introduction}

Simulating user behavior is central to the development of human-centered AI assistants. User simulators provide scalable environments in which AI assistants can be trained \citep{wu2025collabllm, zhou2024archer} and evaluated \citep{yao2024taubenchbenchmarktoolagentuserinteraction, ICLR2024_b3075b88} on the kinds of interactions they encounter in the real world. Yet current simulators struggle to faithfully simulate diverse users and instead demonstrate homogeneous behaviors that tend to resemble those of a generic user \citep{mehri2026measuring, xiao2026chameleonslimitinvestigatingpersona}. Reliance on these simulators exposes assistants to a narrow range of behaviors during development, leading to poor generalization to users that are underrepresented in simulation \citep{liu-etal-2023-one}. It is therefore essential that user simulators capture this diversity and reproduce the behaviors of individual users with high fidelity.

To simulate a particular user, a simulator is conditioned on a user representation and then used to generate responses. A user representation is a natural-language description intended to capture the information needed to reproduce a user's behavior. When relevant information is missing, simulators fill in the gaps with assumptions about how the user might behave, often resulting in generic behaviors rather than the particular user's behavior. Existing works that improve the simulation pipeline primarily focus on optimizing the user simulator to generate more realistic responses, while representation generation receives comparatively little attention and is typically done with an ad hoc prompt and a general-purpose large language model (LLM) \citep{naous2026flipping, wu2026humanlm, zhou2026odyssim}. In this work, we identify user representations as an underexplored bottleneck in user simulation, and optimize the representation generator for downstream simulation fidelity while keeping the user simulators fixed.

The quality of a user representation becomes apparent in hindsight through the behavior it induces in a simulator. We operationalize this with a conversation reconstruction task: we generate a user representation from a user--assistant conversation, condition a fixed simulator on it, and have the simulator interact with an assistant to produce a simulated conversation. We evaluate the representation by measuring how faithfully the simulator reproduces the user behavior from the original conversation. Comparing the two conversations also yields \emph{hindsight feedback}, a description of where the representation falls short and how it can be improved. The representation generator can be conditioned on this feedback to generate an improved representation. Collecting hindsight feedback, however, requires expensive multi-turn simulations. We therefore introduce \emph{hindsight distillation}, an online training procedure in which we distill feedback-conditioned improvements into the representation generator. At each training step, we sample a representation, improve it with hindsight feedback, and then use it as a supervised target. This teaches the representation generator to produce better representations directly from the original conversation.

We present \texttt{UserRepGen-4B}, a representation generator trained with hindsight distillation, and evaluate it on 700 conversations spanning seven datasets of user--AI interactions. \texttt{UserRepGen-4B} generates representations that enable more faithful simulation than those generated by proprietary models such as GPT-5.5, without requiring hindsight feedback at inference time. Improving representations with hindsight feedback also improves simulation fidelity at test time, with further gains when repeated for multiple iterations. Beyond conversation reconstruction, the learned representations improve simulation fidelity for unseen tasks, produce less homogeneous simulations that better preserve the population-level diversity of real users, and capture user-specific behaviors that better distinguish users across conversations.

The main contributions of our work can be summarized as follows:
\begin{itemize}[leftmargin=12pt,topsep=0pt]
\itemsep -0.5ex
\item \textbf{User representations and conversation reconstruction.} We identify user representations as an underexplored bottleneck in user simulation and formulate conversation reconstruction as a task for evaluating and training representation generators (\S~\ref{sec:task-description}).
\item \textbf{Structured user representations.} We structure a user representation into separate task and behavior descriptions, enabling a user's behavior description to be paired with unseen tasks or different behavior descriptions to be paired with a fixed task (\S~\ref{sec:task-description}).
\item \textbf{Hindsight distillation.} We introduce \emph{hindsight distillation}, an online training procedure in which we train a representation generator using \emph{hindsight feedback}, which describes how to improve user representations by comparing real and simulated user behavior. (\S~\ref{sec:hindsight-feedback} \& \S~\ref{sec:hindsight-distillation}).
\item \textbf{\texttt{UserRepGen-4B}.} We present \texttt{UserRepGen-4B}, a representation generator trained with hindsight distillation, whose representations enable more faithful simulation than those generated by proprietary models such as GPT-5.5 (\S~\ref{sec:hindsight-distillation-results}).  Improving representations with hindsight feedback also improves fidelity at test time, and can be applied for multiple iterations (\S~\ref{sec:hindsight-feedback-results}).
\item \textbf{Characterizing learned representations.} We present a systematic analysis of how user representations change after being optimized for simulation fidelity (\S~\ref{sec:analyzing-user-reps}).
\item \textbf{Broader utility of learned representations.} Beyond conversation reconstruction, the learned representations improve simulation fidelity on unseen tasks (\S~\ref{sec:generalization}), produce less homogeneous simulations that preserve the population-level diversity of real users (\S~\ref{sec:population-level}), and better distinguish users across conversations (\S~\ref{sec:user-reid}).
\end{itemize}
\section{Related Work}

\paragraph{User simulation.} LLMs have emerged as promising tools for simulating human behavior \citep{10.1145/3586183.3606763, yao2024taubenchbenchmarktoolagentuserinteraction, anthis2025position, mehri2026multisessioncollablearninguserpreferences}. Yet prompting general-purpose LLMs has revealed several limitations, motivating efforts to train specialized simulators with objectives such as profile consistency, goal alignment, latent-state alignment, and realism \citep{wang-etal-2025-know, mehri2026goal, naous2026flipping, wu2026humanlm, zhou2026odyssim, wang2026learningusersimulatorsturing}. Most recently, studies have documented behavioral homogenization among simulated users and the resulting mismatch with real user populations \citep{mehri2026measuring, xiao2026chameleonslimitinvestigatingpersona}. Our work addresses this problem by conditioning simulators on learned user representations that improve simulation fidelity.

\paragraph{User representation.} Early works introduce personas to enable assistants to generate responses personalized to and consistent with different individuals \citep{li-etal-2016-persona, zhang-etal-2018-personalizing}. User simulation has explored several forms of user representation, including describing user goals \citep{schatzmann-etal-2007-agenda, naous2026flipping}, demographic or biographical attributes \citep{Argyle_Busby_Fulda_Gubler_Rytting_Wingate_2023}, interview transcripts \citep{2024arXiv241110109P}, and learned vector representations \citep{li-etal-2016-persona, 10.1145/3701716.3715463, chen2025personavectorsmonitoringcontrolling}. User representations are central to simulation reliability, directly shaping the behaviors that simulators produce \citep{NEURIPS2025_5fc150b7}. To improve user representations, recent work has explored reducing underspecification \citep{lin2026the}, scaling generation \citep{ge2025scalingsyntheticdatacreation}, generating diverse populations of personas \citep{chopra2026cooperativesimulatorsgeneratingrealistic}, and matching the distribution of behaviors in a reference population \citep{liu2026individualpersonasaligningsynthetic}. We learn to generate user representations that capture the user-specific information necessary for faithful simulation.

\paragraph{Learning from privileged information.} Recent work incorporates environmental feedback into model training through reinforcement learning and self-distillation \citep{song2026expanding, shi2026experiential, sun2026reinforcinghumanbehaviorsimulation}, often by distilling from a teacher conditioned on privileged information \citep{zhao2026selfdistilled, hubotter2026reinforcement, mendes2026making}. For interactive agents, privileged information is obtained from multi-turn rollouts and used to improve the agent \citep{buening2026aligninglanguagemodelsuser, wang2026skillsdskillconditionedselfdistillationmultiturn, wang2026hindsighthintdistillationscaffolded, yeo2026hintsdtargetedhindsightselfdistillation, li2026distillselectivehindsightdistillation, liu2026herohindsightenhancedreflectionenvironment}. In contrast, our setting uses the simulator behavior induced by a representation in a multi-turn conversation as privileged information for improving the representation generator.
\section{Task Description}
\label{sec:task-description}

Let $\mathcal{C} = (u_1, a_1, \ldots, u_n, a_n)$ be a conversation between a real user and an assistant, where $u_i$ and $a_i$ denote the user and assistant utterances at turn $i$, respectively. The goal of user simulation is to generate responses that reproduce the behavior exhibited by the user in $\mathcal{C}$. A typical pipeline consists of two stages. First, a representation generator $G_\theta$ generates a user representation $r$ from $\mathcal{C}$, which is a natural language description that captures the information needed to simulate the user. Second, an LLM-based user simulator $U_\phi$ is conditioned on $r$ and interacts with an assistant to generate a simulated conversation $\tilde{\mathcal{C}} = (\tilde{u}_1, \tilde{a}_1, \ldots, \tilde{u}_m, \tilde{a}_m)$. 

Recent work has largely focused on the second stage, optimizing the user simulator $U_\phi$ to generate responses that resemble real humans \citep{naous2026flipping, wu2026humanlm, zhou2026odyssim}. By contrast, representation generation has received comparatively little attention and is typically performed using a general-purpose LLM and an ad hoc prompt. The representations may fail to capture all relevant information about a user's behavior, which can cause unwanted behaviors from $U_\phi$. For example, the simulator may fill in the gaps with assumptions about the user, and demonstrate generic behaviors rather than those of the particular user in $\mathcal{C}$. This motivates treating representation generation as its own learning problem and optimizing $G_\theta$ for downstream simulation fidelity.

\paragraph{User Representations.} One approach for representing users is through personas that describe demographic or biographical attributes \citep{anthis2025position}, such as age, occupation, location, experiences, interests, personality, or background \citep{zhang-etal-2018-personalizing, ge2025scalingsyntheticdatacreation}. These attributes provide only indirect cues about user behavior, leaving the simulator to fill in the gaps with potentially unsupported or stereotypical assumptions. Another approach represents users through an intent that describes their overall goal without specifying any behaviors \citep{naous2026flipping}. This may suffice for simulating a generic user pursuing a given task, but it does not capture the unique behaviors of individual users, which can result in homogeneous simulations that fail to reflect the diversity of real users \citep{mehri2026measuring, xiao2026chameleonslimitinvestigatingpersona}.

In this work, we introduce a structured user representation $r=(t,b)$ that separates the user's conversation-specific task from their behavioral patterns that persist across conversations. The task $t$ describes what the user is trying to accomplish in $\mathcal{C}$, and behavior $b$ describes the user's general behavior, such as how they formulate and specify requests, respond to the assistant, and provide feedback. This structure enables composing representations, either by simulating diverse user behaviors on a fixed task, or in simulating multiple tasks with a fixed user behavior.

\paragraph{Conversation Reconstruction.} We train and evaluate representation generators through a conversation reconstruction task. Given a conversation $\mathcal{C}$, we measure how well a user representation $r$ enables user simulators to generate conversations that preserve the user behavior observed in $\mathcal{C}$.

We formalize this task with a simulation environment $e=(U_e,A_e,J_e)$, consisting of a fixed user simulator $U_e$, assistant $A_e$, and judge $J_e$. For each simulation environment, the simulator $U_e$ is conditioned on $r$ and interacts with the assistant $A_e$ to produce a simulated conversation $\tilde{\mathcal{C}}_e = (\tilde{u}_1, \tilde{a}_1, \ldots, \tilde{u}_m, \tilde{a}_m)$. The interaction ends when $U_e$ is satisfied and emits a termination signal, or when the conversation reaches a maximum length. The judge $J_e$ compares $\tilde{\mathcal{C}}_e$ with $\mathcal{C}$ and produces a scalar score $s_e \in \mathbb{R}$ that measures simulation fidelity across multiple dimensions (see Appendix~\ref{appendix:prompts}). To account for differences in simulator behavior caused by different assistant responses, $J_e$ evaluates fidelity based on the given interaction context rather than requiring an exact reconstruction. 

Our goal is to train $G_\theta$ to generate representations that maximize fidelity across fixed simulators.
\section{Hindsight Feedback}
\label{sec:hindsight-feedback}

A user representation may appear reasonable without capturing all the relevant information needed to reproduce a user's behavior. Its quality becomes apparent in hindsight, after it is used to simulate the user. The behaviors that the simulator reproduces faithfully indicate what the representation captures well, whereas the behaviors it reproduces poorly or fails to reproduce reveal what the representation does not capture well. We describe these in natural language with \emph{hindsight feedback}.

Given a conversation $\mathcal{C}$ and an initial user representation $r^{(0)}$, we collect hindsight feedback for the representation using an ensemble of simulation environments $\mathcal{E}=\{e_k=(U_k,A_k,J_k)\}_{k=1}^{K}$, each built on a different underlying model. For each environment $e_k$, we roll out a simulated conversation:
\begin{equation}
    \tilde{\mathcal{C}}_k
    \sim
    \operatorname{rollout}\!\left(
        U_k(\cdot \mid r^{(0)}),A_k
    \right).
\end{equation}
The judge $J_k$ then compares the user behavior in $\tilde{\mathcal{C}}_k$ with that in $\mathcal{C}$ and generates hindsight feedback $f_k$. We collect the feedback from all simulation environments into $f=(f_1,\ldots,f_K)$, where
\begin{equation}
    f_k
    \sim
    J_k\!\left(
        \cdot \mid \mathcal{C},\tilde{\mathcal{C}}_k
    \right),
    \qquad k=1,\ldots,K.
\end{equation}

\paragraph{Improving user representations with hindsight feedback.} We leverage the in-context learning capabilities of LLMs to improve a user representation. We generate an improved representation by conditioning $G_\theta$ on the conversation $\mathcal{C}$, the current representation $r$, and the hindsight feedback $f$. We repeat this improvement for $I$ iterations, as each iteration may expose previously unseen failures or reveal that earlier improvements were insufficient. At each round $i$, we collect feedback $f^{(i)}$ for $r^{(i)}$ and generate:
\begin{equation}
    r^{(i+1)}
    \sim
    G_\theta\!\left(
        \cdot \mid
        \mathcal{C},r^{(i)},f^{(i)}
    \right),
\end{equation}
The final user representation is denoted by $r^{(I)}$.

\section{Hindsight Distillation}
\label{sec:hindsight-distillation}

\begin{wrapfigure}{r}{0.45\linewidth}
    \centering
    \vspace{-2em}
    \includegraphics[width=\linewidth]{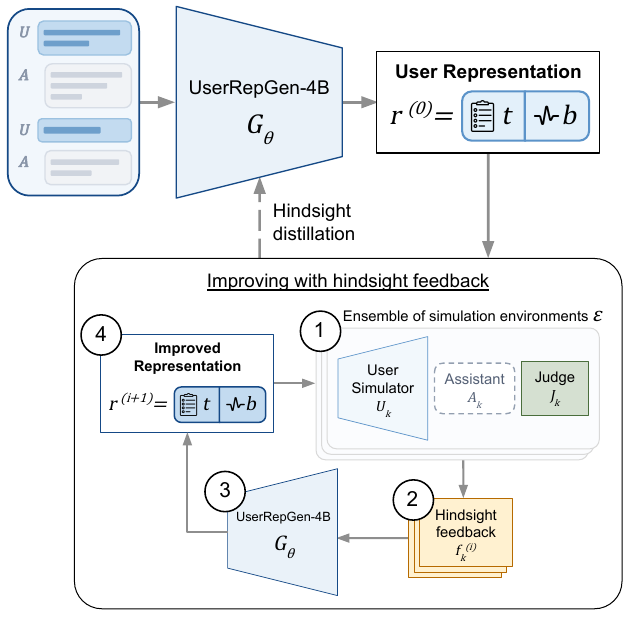}
    \vspace{-2em}
    \caption{Hindsight distillation for $G_\theta$. A representation is sampled, and is improved with $I$ rounds of hindsight feedback to get $r^{(I)}$, which serves as a training target.}
    \vspace{-2em}
    \label{fig:method}
\end{wrapfigure}

Hindsight feedback enables $G_\theta$ to improve a user representation, but obtaining this feedback requires simulating and evaluating a multi-turn conversation. To avoid this during inference, we distill the feedback-conditioned improvements into $G_\theta$, enabling it to generate improved representations directly from $\mathcal{C}$.

At training step $n$, the current policy samples an initial user representation for a conversation $\mathcal{C}$
\begin{equation}
    r^{(0)}
    \sim
    G_{\theta_n}(\cdot \mid \mathcal{C}).
\end{equation}
We apply $I$ rounds of hindsight feedback to obtain $r^{(I)}=(r_1^{(I)},\ldots,r_L^{(I)})$, which we use as a training target under the negative log-likelihood objective:
\begin{equation}
    \mathcal{L}(\theta_n)
    =
    -\frac{1}{L}
    \sum_{j=1}^{L}
    \log
    G_{\theta_n}
    \left(
        r_j^{(I)}
        \mid
        \mathcal{C},
        r_{<j}^{(I)}
    \right).
\end{equation}
This distills hindsight feedback-conditioned improvements into $G_\theta(\cdot \mid \mathcal{C})$, teaching it to generate better representations without the hindsight feedback. The complete training procedure is depicted in Figure~\ref{fig:method} and Algorithm~\ref{alg:hindsight-distillation} in Appendix~\ref{appendix:algm}.
\section{Learning User Representations}
\label{sec:experiments}

\subsection{Experimental Setup}
\label{sec:experimental-setup}

\paragraph{Datasets.} We use five datasets of real-world conversations with AI assistants: WildChat \citep{zhao2024wildchat}, LMSYS-Chat-1M \citep{ICLR2024_5f9bfdfe}, PRISM \citep{NEURIPS2024_be2e1b68}, SWE-Chat \citep{baumann2026swechatcodingagentinteractions}, and ThoughtTrace \citep{jin2026thoughttraceunderstandinguserthoughts}. We additionally use two datasets of simulated conversations: WildChat Coding and WildChat Writing \citep{mehri2026measuring}. For each dataset, we apply a preprocessing stage that removes near-duplicate conversations based on n-gram overlap, and retains safe English conversations with at least four user turns.

\paragraph{Training.} We use a fixed set of 3,200 conversations, approximately 500 conversations from each dataset and 200 from ThoughtTrace. Hindsight feedback is generated using an ensemble of two simulation environments based on Qwen3.5-4B \citep{qwen35blog} and Gemma3-4B \citep{gemmateam2025gemma3technicalreport}. Within each environment, the corresponding model serves as the user simulator, assistant and judge. During training, the judges compare only user utterances from the real and simulated conversations, which we found to be a reliable proxy for the full-conversation evaluation when using smaller and less capable judge models. Additionally, we stabilize training by using the initial policy as our teacher policy \citep{zhao2026selfdistilled} and to generate improved user representations.

\paragraph{Evaluation.} We sample 100 conversations from each dataset, yielding 700 conversations in total. We follow the evaluation process described in Section~\ref{sec:task-description}, using GPT-5.4 mini as the user simulator, assistant, and judge. We specifically use a different model than training to ensure a more reliable and unbiased evaluation that tests whether the learned representations generalize to unseen simulation environments. The judge compares user behavior in simulated conversations with that in the corresponding original conversation and assigns an overall fidelity score from 1 to 10. We validate the judge with a human preference study in Appendix~\ref{appendix:judge-validation}. Lastly, simulators sometimes demonstrated failures in staying aligned with their user representations \citep{mehri2026goal}. To account for this, we simulate three conversations for each representation, and report \emph{best-of-1} and \emph{best-of-3} fidelity. All prompts are provided in Appendix~\ref{appendix:prompts} and hyperparameters are reported in Appendix~\ref{appendix:hyperparams}.

\subsection{Test-Time Improvement with Hindsight Feedback}
\label{sec:hindsight-feedback-results}

\begin{wrapfigure}{r}{0.35\linewidth}
    \centering
    \vspace{-1em}
    \includegraphics[width=\linewidth]{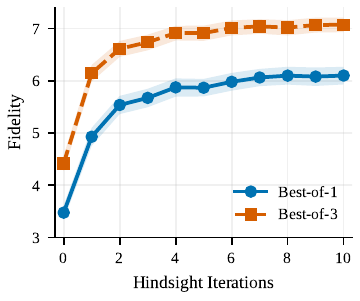}
    \vspace{-2em}
    \caption{Simulation fidelity across iterations of hindsight feedback. Shaded regions show 95\% bootstrap confidence intervals over conversations.}
    \vspace{-3em}
    \label{fig:iterative-improvement}
\end{wrapfigure}

We investigate whether hindsight feedback improves user representations directly at test time. For each evaluation conversation, we generate an initial user representation $r^{(0)}$ using Qwen3.5-4B. Using the ensemble $\mathcal{E}$ of Qwen3.5-4B and Gemma3-4B environments, we apply $I=10$ rounds of improvement using hindsight feedback, as described in Section~\ref{sec:hindsight-feedback}. We evaluate representations after every iteration (Figure~\ref{fig:iterative-improvement}).

Hindsight feedback helps generate better user representations at test-time, which leads to higher simulation fidelity scores. After one round of feedback, best-of-1 fidelity increases from 3.41 to 4.93, and best-of-3 increases from 4.42 to 6.14. Performance continues to improve over subsequent rounds, with best-of-1 reaching 6.10 after ten iterations and best-of-3 peaking at 7.07 at nine. Most gains occur in the first two rounds, and performance largely plateaus after six rounds.

\subsection{Hindsight Distillation}
\label{sec:hindsight-distillation-results}

We then distill hindsight feedback into the representation generator to generate better representations directly from a conversation. Unless otherwise stated, hindsight feedback is collected using the ensemble of simulation environments described in Section~\ref{sec:experimental-setup}. We explore four training approaches:
\begin{itemize}[leftmargin=12pt,topsep=0pt,itemsep=0pt]
    \item \textbf{Supervised Fine-Tuning (SFT):} For each conversation $\mathcal{C}$ in the training set, the base policy samples an initial user representation $r^{(0)}\sim G_{\theta_0}(\cdot\mid\mathcal{C})$. We collect hindsight feedback and generate an improved user representation $r^{(1)}$. We then train the representation generator to predict $r^{(1)}$ directly from $\mathcal{C}$ using the standard negative log-likelihood objective.
    \item \textbf{Group Relative Policy Optimization (GRPO)  \citep{shao2024deepseekmathpushinglimitsmathematical}:} For each conversation $\mathcal{C}$, the current policy samples a group of $n$ user representations $\{r_i\}_{i=1}^{n}$. We assign each group to one of the simulation environments $e=(U_e,A_e,J_e)$ and roll out a simulated conversation $\tilde{\mathcal{C}}_{e,i} \sim \operatorname{rollout}\!\left(U_e(\cdot \mid r_i),A_e \right)$ for each representation $r_i$. To compute the reward, the judge assigns a scalar score $s_i=J_e(\mathcal{C}, \tilde{\mathcal{C}}_{e,i})$ measuring simulation fidelity. We compute group-relative advantages and update the policy using the clipped GRPO objective.
    \item \textbf{On-Policy Self-Distillation (OPSD) \citep{zhao2026selfdistilled}:} The current policy samples an initial user representation $r^{(0)}\sim G_{\theta}(\cdot\mid\mathcal{C})$. Hindsight feedback $f$ is collected, and the teacher policy is conditioned on the privileged context $(\mathcal{C},r^{(0)},f)$, while the student policy is conditioned only on $\mathcal{C}$. We update the student policy by minimizing the token-level forward KL divergence from the teacher distribution to the student distribution along $r^{(0)}$.
    \item \textbf{Online Supervised Fine-Tuning (OSFT):} As described in Section~\ref{sec:hindsight-distillation}, the current policy samples an initial user representation $r^{(0)}\sim G_{\theta}(\cdot\mid\mathcal{C})$ and receives hindsight feedback $f$. Rather than using token-level supervision from the teacher policy, we condition the policy on $(\mathcal{C},r^{(0)},f)$ to generate an improved representation $r^{(1)}$. The policy is updated to predict $r^{(1)}$ directly from $\mathcal{C}$ using the standard negative log-likelihood objective.
\end{itemize}

Results for training Qwen3.5-4B using the different approaches are reported in Table~\ref{tab:hindsight-distillation-results}. OSFT performs best, achieving 4.95 best-of-1 and 6.02 best-of-3. We refer to this model as \texttt{UserRepGen-4B}. OPSD achieves 4.46 best-of-1 and 5.60 best-of-3. Like OSFT, OPSD samples user representations from the current policy and collects hindsight feedback online. However, OPSD distills this feedback through dense token-level supervision, whereas OSFT generates an improved representation and uses it as a supervised target. The stronger performance of OSFT suggests that explicitly supervising on an improved representation is more effective for this task. SFT similarly trains on improved user representations using negative log-likelihood, but generates its targets only once from the base policy rather than continually doing so using the latest policy. It scores 4.64 best-of-1 and 5.68 best-of-3, therefore supporting prior findings and showing that an online approach for collecting feedback and improving representations is more effective \citep{pmlr-v15-ross11a, ICLR2024_5be69a58, pmlr-v235-xiong24a, shao2024deepseekmathpushinglimitsmathematical}.

\definecolor{darkgreen}{rgb}{0.0, 0.5, 0.0}
\providecommand{\ccol}[1]{\cellcolor[HTML]{#1}}

\begin{wraptable}{r}{0.4\textwidth}
\centering
\vspace{-2em}
\caption{Simulation fidelity across user representation generators. We report Best-of-1 and Best-of-3 scores. Subscripts indicate absolute improvements over the Qwen3.5-4B, and darker blue indicates stronger scores.}
\label{tab:hindsight-distillation-results}
\resizebox{\linewidth}{!}{
\begin{tabular}{l|cc}
\toprule
\textbf{Model} & \textbf{Best-of-1} & \textbf{Best-of-3} \\
\midrule
Qwen3.5-4B
    & \ccol{F6FBFE} 3.41
    & \ccol{F6FBFE} 4.42 \\
\quad + GRPO
    & \ccol{EBF6FD} 3.57$_{\uparrow 0.16}$
    & \ccol{EBF6FD} 4.59$_{\uparrow 0.17}$ \\
\quad + SFT
    & \ccol{A1D5F3} 4.64$_{\uparrow 1.23}$
    & \ccol{A2D6F3} 5.68$_{\uparrow 1.26}$ \\
\quad + OPSD
    & \ccol{ADDBF4} 4.46$_{\uparrow 1.05}$
    & \ccol{A7D8F4} 5.60$_{\uparrow 1.18}$ \\
\quad + OSFT
    & \ccol{8BCCF0} 4.95$_{\uparrow 1.54}$
    & \ccol{8BCCF0} 6.02$_{\uparrow 1.60}$ \\

\cmidrule(lr){1-3}
Qwen3.5-9B
    & \ccol{E0F2FB} 3.72$_{\uparrow 0.31}$
    & \ccol{E5F3FC} 4.68$_{\uparrow 0.26}$ \\
Qwen3.5-27B
    & \ccol{DCEFFB} 3.79$_{\uparrow 0.38}$
    & \ccol{DBEFFA} 4.83$_{\uparrow 0.41}$ \\
Qwen3.5-122B-A10B
    & \ccol{EEF7FD} 3.53$_{\uparrow 0.12}$
    & \ccol{EDF7FD} 4.55$_{\uparrow 0.13}$ \\
GPT-5.4 mini
    & \ccol{E3F3FC} 3.68$_{\uparrow 0.27}$
    & \ccol{E3F2FB} 4.71$_{\uparrow 0.29}$ \\
GPT-5.5
    & \ccol{D4ECFA} 3.90$_{\uparrow 0.49}$
    & \ccol{D6EDFA} 4.90$_{\uparrow 0.48}$ \\
\bottomrule
\end{tabular}
}
\vspace{-3em}
\end{wraptable}

GRPO achieves 3.57 best-of-1 and 4.59 best-of-3, yielding only modest improvements over the base model. Two factors may help explain this result. First, scalar rewards provide limited information about what a user representation omitted or how it should be improved \citep{song2026expanding, hubotter2026reinforcement, zhao2026selfdistilled, sun2026reinforcinghumanbehaviorsimulation}. This limitation is especially costly in our setting because computing each reward requires an expensive multi-turn simulation. Second, we observe low within-group reward diversity during training, with a median reward variance of 0.029 (mean 0.048). This can reduce the learning signal, and when all samples receive identical rewards, their advantages are zero and have no gradient signal  \citep{hubotter2026reinforcement,zhao2026selfdistilled}.

\paragraph{General-purpose LLMs struggle to generate user representations.} We also investigate whether stronger general-purpose LLMs generate better user representations without training and report results in Table~\ref{tab:hindsight-distillation-results}. Despite substantial differences in model scale and general capability, all models perform within a similar range. GPT-5.5, the strongest off-the-shelf model, outperforms Qwen3.5-4B by only 0.49 best-of-1 and 0.48 best-of-3 points. In contrast, \texttt{UserRepGen-4B} surpasses GPT-5.5 and raises scores by 1.05 and 1.12 points. These results suggest that user representation generation depends less on general model capability and more on explicitly learning what information is important for downstream simulation fidelity.

\subsection{How Does Training Change User Representations?}
\label{sec:analyzing-user-reps}

\begin{wraptable}{r}{0.47\textwidth}
\centering
\vspace{-2em}
    \caption{Quantitative analysis of user representations. We report the average word counts of the task, behavior and complete user representations, as well as ROUGE-L precision against the source conversation.}
\label{tab:quantitative}
\resizebox{\linewidth}{!}{
\begin{tabular}{l|rrrr}
\toprule
\textbf{Model}
    & \textbf{Task}
    & \textbf{Behavior}
    & \textbf{Total}
    & \textbf{R-L (\%)} \\
\midrule
Qwen3.5-4B
    & 61
    & 101
    & 163
    & 27.2 \\
\quad + GRPO
    & 61
    & 101
    & 162
    & 27.8 \\
\quad + SFT
    & 58
    & 126
    & 185
    & 25.8 \\
\quad + OPSD
    & 72
    & 111
    & 183
    & 25.3 \\
\quad + OSFT
    & 62
    & 148
    & 210
    & 23.9 \\
\bottomrule
\end{tabular}
}
\vspace{-1em}
\end{wraptable}

We analyze how user representations change when optimized for downstream simulation fidelity. Quantitatively, we measure their average length and ROUGE-L precision \citep{lin2004rouge}, which measures lexical overlap with the source conversation. Table~\ref{tab:quantitative} shows that the representation lengths stay within the same general range, with slight increase in the behavior descriptions. ROUGE-L precision decreases slightly, indicating that improvements in fidelity do not come from simply copying from the source conversation.

We next qualitatively analyze the task and behavior descriptions. After training, task descriptions become more standardized and comprehensive, more consistently capturing constraints introduced in later turns, shifts in direction, and multiple objectives pursued within a single conversation. Behavior descriptions become more specific and actionable. Rather than relying on broad characterizations such as ``goal-oriented'' or ``directive'', they describe observable patterns that a simulator can reproduce. They also use more negative and contrastive instructions, such as specifying that a user does not provide all constraints upfront but instead introduces them incrementally. Lastly, the descriptions capture dimensions that were previously overlooked, including how users respond to the assistant, formulate requests, communicate, provide context, and engage across turns. Appendix~\ref{appendix:rep-analysis} uses GPT-5.4 mini to analyze these observations and provides qualitative examples.

\providecommand{\conditionlabel}[1]{\textcircled{\scriptsize #1}}

\section{Broader Utility of Learned Representations}
\label{sec:discussion}

Conversation reconstruction provides an objective for learning and evaluating user representations based on simulation fidelity. We next demonstrate their broader utility beyond reconstruction.

\subsection{Simulating Users in New Scenarios}
\label{sec:generalization}

One use case for user simulators is to use evidence from a user's prior interactions to simulate how they might behave in new scenarios. Our primary evaluation, however, evaluates a representation through reconstruction of the same conversation from which it was generated. We therefore evaluate how faithfully a behavior description enables user simulation for unseen tasks.

We select all 147 users from ThoughtTrace who have at least two conversations and do not appear in our training set. For each user, let $\mathcal{C}$ denote the target conversation and $\mathcal{C}'$ another conversation from the same user. We generate representations $r=(t,b) \sim G_\theta(\mathcal{C})$ and $r'=(t',b') \sim G_\theta(\mathcal{C}')$. We follow Section~\ref{sec:experimental-setup} and use GPT-5.4 mini as the user simulator, assistant and judge. We condition it under four configurations while keeping task description $t$ fixed:
\begin{enumerate}[leftmargin=12pt,topsep=0pt,itemsep=0pt]
    \item[\conditionlabel{A}]\textbf{Same task only:} Task $t$ from $\mathcal{C}$ without a behavior description.
    \item[\conditionlabel{B}] \textbf{Same task, different user, different conversation:} Task $t$ and behavior $\bar{b}$ from a randomly selected conversation by a different user.
    \item[\conditionlabel{C}] \textbf{Same task, same user, different conversation:} Task $t$ and behavior $b'$ from the same user's other conversation $\mathcal{C}'$.
    \item[\conditionlabel{D}] \textbf{Same task, same user, same conversation:} Task $t$ and behavior $b$ from $\mathcal{C}$.
\end{enumerate}


\begin{figure*}[h]
    \centering
    \includegraphics[width=0.9\textwidth]{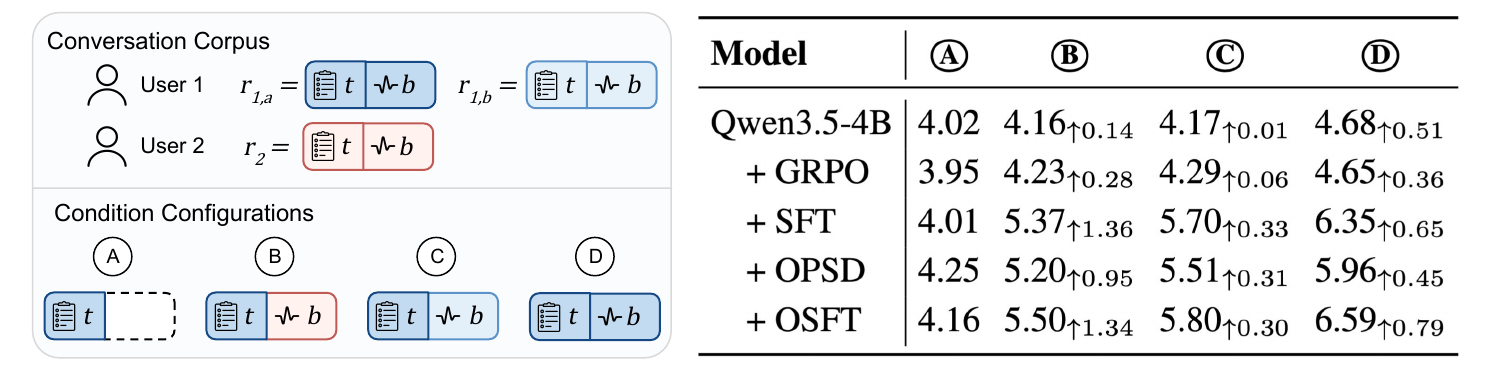}
    \caption{Simulation fidelity when conditioned under configurations \conditionlabel{A}, \conditionlabel{B}, \conditionlabel{C}, and \conditionlabel{D}. Each cell reports Best-of-3 scores. Subscripts indicate absolute improvement relative to the condition immediately to the left.}
    \label{fig:conditions}
\end{figure*}

For each configuration, the simulator interacts with the assistant to generate a simulated conversation $\tilde{\mathcal{C}}$. The judge compares the user behavior in $\tilde{\mathcal{C}}$ with that in $\mathcal{C}$ and assigns a fidelity score. We generate three rollouts per configuration and report best-of-3 fidelity in Figure~\ref{fig:conditions}.

\paragraph{\conditionlabel{B} $\gg$ \conditionlabel{A}: Behavior descriptions encourage realistic simulation.} Adding a behavior description improves fidelity even when it is generated from a different user's conversation. The descriptions provide concrete guidance, such as how a user formulates requests or responds to the assistant. This helps the simulator avoid defaulting to assistant-like behaviors and be more representative of real users in general, despite not identifying the behavior of the particular user being simulated. The larger gain for trained representation generators indicates more effective behavior descriptions.

\paragraph{\conditionlabel{C} $>$ \conditionlabel{B}: Behavior descriptions improve simulation for unseen tasks.}
Condition \conditionlabel{C} tests whether behavior extracted from one of a user's conversations enables faithful simulation of the same user in another conversation with an unseen task. \conditionlabel{C} outperforms \conditionlabel{B} for every trained representation generator, providing evidence that the descriptions capture user-specific behaviors that generalize to new scenarios. The improvements are modest, however, and remain well below \conditionlabel{D}, an upper bound using behavior extracted directly from the target conversation.


\subsection{Preserving Behavioral Diversity at the Population Level}
\label{sec:population-level}

We next evaluate user representations at the population level, asking whether the learned representations produce less homogeneous simulations that better preserve the behavioral diversity of the original conversations. We measure the distributional gap between original and simulated user behaviors using the approach proposed by \citet{mehri2026measuring}: (1) Qwen3.5-122B-A10B-FP8 \citep{qwen35blog} describes the user behavior in each original and simulated conversation along six dimensions; (2) Qwen3-Embedding-8B \citep{qwen3embedding} embeds the resulting descriptions; (3) the embeddings are quantized using $k$-means into $k=500$ clusters; (4) the cluster assignments for each conversation are used to construct discrete distributions for the original and simulated user behaviors; and (5) the two distributions are  compared using Jensen--Shannon divergence ($\mathrm{JS}$). See Appendix~\ref{appendix:distributional-gap} for details.

\definecolor{darkgreen}{rgb}{0.0, 0.5, 0.0}
\definecolor{okabe_blue}{HTML}{56B4E9}
\providecommand{\ccol}[1]{\cellcolor[HTML]{#1}}

\begin{wraptable}{r}{0.5\textwidth}
\centering
\vspace{-2em}
\small
\setlength{\tabcolsep}{7pt}
\caption{Distributional gap between original and simulated user behaviors for our different representation generators when the simulator is conditioned on the task description alone or both the task and behavior descriptions. We report ($\mathrm{JS}$) divergence and Self-BLEU (SB). Darker shades of blue and lower values indicate better performance.}
\label{tab:distributional-gap}

\resizebox{\linewidth}{!}{%
\begin{tabular}{l|cc|cc}
\toprule
& \multicolumn{2}{c|}{\textbf{Task Only}}
& \multicolumn{2}{c}{\textbf{Task + Behavior}} \\
\cmidrule(lr){2-3}
\cmidrule(lr){4-5}
\textbf{Model}
& $\mathrm{JS}$\,$\downarrow$
& SB\,$\downarrow$
& $\mathrm{JS}$\,$\downarrow$
& SB\,$\downarrow$ \\
\midrule
Qwen3.5-4B
  & \ccol{FFFFFF} 0.389
  & \ccol{EFF8FD} 0.257
  & \ccol{96D0F1} 0.315
  & \ccol{A8D8F4} 0.227 \\
  \quad + GRPO
  & \ccol{D0EAF9} 0.356
  & \ccol{ECF7FD} 0.256
  & \ccol{B7DFF6} 0.338
  & \ccol{91CEF1} 0.217 \\
  \quad + SFT
  & \ccol{CCE8F8} 0.353
  & \ccol{FFFFFF} 0.264
  & \ccol{56B4E9} 0.270
  & \ccol{5FB8EA} 0.196 \\
  \quad + OPSD
  & \ccol{A6D7F3} 0.326
  & \ccol{C4E5F7} 0.239
  & \ccol{8CCCF0} 0.308
  & \ccol{6DBEEC} 0.202 \\
  \quad + OSFT
  & \ccol{CDE9F9} 0.354
  & \ccol{D5ECFA} 0.246
  & \ccol{5AB6EA} 0.273
  & \ccol{56B4E9} 0.192 \\
\bottomrule
\end{tabular}
}
\vspace{-1.5em}
\end{wraptable}

For each representation generator, we generate a user representation for each of the 700 evaluation conversations. Using GPT-5.4 mini as both the user simulator and assistant, we produce a corresponding simulated conversation and measure the distributional gap. We additionally report Self-BLEU \citep{10.1145/3209978.3210080}, which measures the lexical overlap between each simulated conversation with the other simulated conversations. We conduct this evaluation under two configurations: conditioning the simulator on only the generated task description and conditioning it on both the task and behavior descriptions. Results are reported in Table~\ref{tab:distributional-gap}.

Across both Task Only and Task + Behavior, the trained representation generators generally achieve lower $\mathrm{JS}$ divergence and Self-BLEU than the base model. Moreover, conditioning the simulator on both the task and behavior descriptions instead of just on the task reduces both metrics for every representation generator. Lower $\mathrm{JS}$ divergence indicates closer alignment between original and simulated distributions of user behavior, and lower Self-BLEU indicates less lexical homogeneity among the simulated conversations. Together, these results show that learned representations better preserve behavioral diversity at the population level.

\subsection{Distinguishing Users Across Conversations}
\label{sec:user-reid}
\definecolor{darkgreen}{rgb}{0.0, 0.5, 0.0}
\definecolor{baselinegray}{HTML}{E6E6E6}

\providecommand{\ccol}[1]{\cellcolor[HTML]{#1}}

\begin{wraptable}{r}{0.45\textwidth}
\centering
\vspace{-2em}
\caption{GPT-5.4 mini accuracy at distinguishing users across conversations. Subscripts indicate performance compared to the full conversation baseline, and darker blue indicates higher accuracy.}
\label{tab:reid}
\resizebox{\linewidth}{!}{
\begin{tabular}{l|c}
\toprule
\textbf{Representation Generator}
  & \textbf{Accuracy (\%)} \\
\midrule
\rowcolor{baselinegray} Full conversation (baseline)
  & 44.22 \\
\midrule
Qwen3.5-4B
  & \ccol{F6FBFE} 38.78$_{\downarrow 5.44}$ \\
\quad + GRPO
  & \ccol{EBF6FD} 41.50$_{\downarrow 2.72}$ \\
\quad + SFT
  & \ccol{DCEFFB} 43.54$_{\downarrow 0.68}$ \\
\quad + OPSD
  & \ccol{91CEF1} 50.34$_{\uparrow 6.12}$ \\
\quad + OSFT
  & \ccol{8BCCF0} 51.02$_{\uparrow 6.80}$ \\
\bottomrule
\end{tabular}
}
\vspace{-2em}
\end{wraptable}

Finally, we consider the task of distinguishing users across conversations. We frame this as a multiple-choice task: given a query and four candidate conversations, the goal is to select the candidate from the same user as the query. We use the same 147 ThoughtTrace users from Section~\ref{sec:generalization}. For each user, one conversation serves as the query and their other conversation serves as the correct candidate. The remaining candidates are conversations from randomly selected other users. We ask GPT-5.4 mini to select the correct candidate and report accuracy. Random selection would yield 25\% accuracy, whereas GPT-5.4 mini achieves 44.22\% accuracy.

We next ask whether user representations can improve performance. For each representation generator, we replace every conversation with its generated behavior description and ask GPT-5.4 mini to solve the same task. Table~\ref{tab:reid} reports the results. Representations generated by the base model yield the lowest accuracy at 38.78\%, 5.44\% under the full conversation baseline. GRPO and SFT narrow the gap, but also remain below the full conversation baseline. OPSD and OSFT achieve the highest accuracies at 50.34\% and 51.02\%, exceeding the full conversation baseline by 6.12\% and 6.80\%. These results demonstrate that the learned representations capture user-specific characteristics that distinguish a user from other users and persist across conversations.

\section{Conclusion}

User representations are a bottleneck in user simulation. A simulator conditioned on a representation can only reproduce the behaviors that the representation captures. When relevant behaviors are missing, the simulator fills in the gaps with assumptions, often producing generic rather than user-specific behavior. We therefore present \texttt{UserRepGen-4B}, a representation generator that we optimize for downstream simulation fidelity using hindsight distillation. Our results show that learned representations enable more faithful simulations of a particular user, improve simulation fidelity for unseen tasks, better preserve the population-level diversity of real users, and better distinguish users across conversations. Simulation fidelity therefore depends not only on the ability to generate realistic user behavior, but also on capturing the relevant information with user representations.

\section*{Acknowledgements}

We would like to thank Matthew Ho, Joseph Suh, Saujas Vaduguru, Kshitish Ghate, Nikhil Sharma, Nishant Balepur, Ethan Mendes, and Xuhui Zhou for feedback at various stages of the projects, as well as researchers in the Deep Learning and AI Interaction and Learning groups at Microsoft Research, including Nebojsa Jojic, Baolin Peng, and Jihoon Tack.

\section*{AI Use Statement}
In this work, generative AI tools were used to help with writing code and implementing methods, obtaining feedback on research methodology and experiments, supporting qualitative analysis of user representations, and assisting with interpreting results. We did not use generative AI tools to develop conceptual frameworks, propose or refine hypotheses, or to directly clean and reformat datasets. Additionally, we used generative AI tools to analyze existing literature, search for information, and improve writing and readability. Writing was limited to expressing the authors' intended content. Mathematical claim formulation, proofs, and translations are not applicable to this work. The use of generative AI tools was to support rather than replace the authors' work. All AI-assisted work was reviewed and verified, and we take responsibility for the final content of this paper, including all text, claims, and artifacts produced with the aid of generative AI.

\section*{Ethics Statement}
\texttt{UserRepGen-4B} is intended to support research on user simulation and the interactive training and evaluation of AI assistants. More faithful simulations may help developers account for a broader range of user behaviors, but also introduce risks of privacy loss, mischaracterization, and misuse. User representations are inferred from a single conversation and may retain sensitive information or make unsupported assumptions about an individual. Our results in Section~\ref{sec:user-reid} show that learned behavior descriptions can help link conversations from the same user, highlighting potential privacy or anonymity considerations. Although \texttt{UserRepGen-4B} enables more faithful simulation of individual users, it should not be used to impersonate them or present simulated interactions as their own. These simulations should not be used to replace real users.

Our training and evaluation use datasets of real English conversations collected through publicly accessible chatbot interfaces, paid participant studies, and agent coding session records in public GitHub repositories. We also use LLM-generated simulated conversations. These sources do not ensure representative coverage of user populations, and we do not confirm whether the observed improvements in simulation fidelity will generalize to other languages and user populations. Furthermore, although our LLM judge was validated against human preferences, we acknowledge that its evaluations may still reflect model biases and inconsistencies.

\section*{Reproducibility Statement}
\texttt{UserRepGen-4B} will be made publicly available. We describe the dataset sources, preprocessing criteria, and training and evaluation setup in Section~\ref{sec:experimental-setup}. Prompts for representation generation, simulation, and evaluation are provided in Appendix~\ref{appendix:prompts}.

\bibliography{iclr2027_conference}

@inproceedings{
naous2026flipping,
title={Flipping the Dialogue: Training and Evaluating User Language Models},
author={Tarek Naous and Philippe Laban and Wei Xu and Jennifer Neville},
booktitle={The Fourteenth International Conference on Learning Representations},
year={2026},
url={https://openreview.net/forum?id=ykSmkVqzn4}
}

@article{wu2026humanlm,
  title={Humanlm: Simulating users with state alignment beats response imitation},
  author={Wu, Shirley and Choi, Evelyn and Khatua, Arpandeep and Wang, Zhanghan and He-Yueya, Joy and Weerasooriya, Tharindu Cyril and Wei, Wei and Yang, Diyi and Leskovec, Jure and Zou, James},
  journal={arXiv preprint arXiv:2603.03303},
  year={2026}
}

@article{zhou2026odyssim,
  title={OdysSim: Building Foundation Models for Human Behavior Simulation},
  author={Zhou, Xuhui and Sun, Weiwei and Du, Weihua and Liu, Jiarui and Sun, Haojia and Ma, Qianou and Wu, Tongshuang and Yang, Yiming and Sap, Maarten},
  journal={arXiv preprint arXiv:2606.14199},
  year={2026}
}

@misc{ge2025scalingsyntheticdatacreation,
      title={Scaling Synthetic Data Creation with 1,000,000,000 Personas}, 
      author={Tao Ge and Xin Chan and Xiaoyang Wang and Dian Yu and Haitao Mi and Dong Yu},
      year={2025},
      eprint={2406.20094},
      archivePrefix={arXiv},
      primaryClass={cs.CL},
      url={https://arxiv.org/abs/2406.20094}, 
}

@inproceedings{zhang-etal-2018-personalizing,
    title = "Personalizing Dialogue Agents: {I} have a dog, do you have pets too?",
    author = "Zhang, Saizheng  and
      Dinan, Emily  and
      Urbanek, Jack  and
      Szlam, Arthur  and
      Kiela, Douwe  and
      Weston, Jason",
    editor = "Gurevych, Iryna  and
      Miyao, Yusuke",
    booktitle = "Proceedings of the 56th Annual Meeting of the Association for Computational Linguistics (Volume 1: Long Papers)",
    month = jul,
    year = "2018",
    address = "Melbourne, Australia",
    publisher = "Association for Computational Linguistics",
    url = "https://aclanthology.org/P18-1205/",
    doi = "10.18653/v1/P18-1205",
    pages = "2204--2213"
}

@inproceedings{zhao2024wildchat,
title={WildChat: 1M Chat{GPT} Interaction Logs in the Wild},
author={Wenting Zhao and Xiang Ren and Jack Hessel and Claire Cardie and Yejin Choi and Yuntian Deng},
booktitle={The Twelfth International Conference on Learning Representations},
year={2024},
url={https://openreview.net/forum?id=Bl8u7ZRlbM}
}

@inproceedings{ICLR2024_5f9bfdfe,
 author = {Zheng, Lianmin and Chiang, Wei-Lin and Sheng, Ying and Li, Tianle and Zhuang, Siyuan and Wu, Zhanghao and Zhuang, Yonghao and Li, Zhuohan and Lin, Zi and Xing, Eric  and Gonzalez, Joseph E and Stoica, Ion and Zhang, Hao},
 booktitle = {International Conference on Learning Representations},
 editor = {B. Kim and Y. Yue and S. Chaudhuri and K. Fragkiadaki and M. Khan and Y. Sun},
 pages = {22225--22257},
 title = {LMSYS-Chat-1M: A Large-Scale Real-World LLM Conversation Dataset},
 url = {https://proceedings.iclr.cc/paper_files/paper/2024/file/5f9bfdfe3685e4ccdbc0e7fb29cccf2a-Paper-Conference.pdf},
 volume = {2024},
 year = {2024}
}

@inproceedings{NEURIPS2024_be2e1b68,
 author = {Kirk, Hannah Rose and Whitefield, Alexander and R\"{o}ttger, Paul and Bean, Andrew and Margatina, Katerina and Ciro, Juan and Mosquera, Rafael and Bartolo, Max and Williams, Adina and He, He and Vidgen, Bertie and Hale, Scott},
 booktitle = {Advances in Neural Information Processing Systems},
 doi = {10.52202/079017-3342},
 editor = {A. Globerson and L. Mackey and D. Belgrave and A. Fan and U. Paquet and J. Tomczak and C. Zhang},
 pages = {105236--105344},
 publisher = {Curran Associates, Inc.},
 title = {The PRISM Alignment Dataset: What Participatory, Representative and Individualised Human Feedback Reveals About the Subjective and Multicultural Alignment of Large Language Models},
 url = {https://proceedings.neurips.cc/paper_files/paper/2024/file/be2e1b68b44f2419e19f6c35a1b8cf35-Paper-Datasets_and_Benchmarks_Track.pdf},
 volume = {37},
 year = {2024}
}

@misc{baumann2026swechatcodingagentinteractions,
      title={SWE-chat: Coding Agent Interactions From Real Users in the Wild}, 
      author={Joachim Baumann and Vishakh Padmakumar and Xiang Li and John Yang and Diyi Yang and Sanmi Koyejo},
      year={2026},
      eprint={2604.20779},
      archivePrefix={arXiv},
      primaryClass={cs.AI},
      url={https://arxiv.org/abs/2604.20779}, 
}

@misc{jin2026thoughttraceunderstandinguserthoughts,
      title={ThoughtTrace: Understanding User Thoughts in Real-World LLM Interactions}, 
      author={Chuanyang Jin and Binze Li and Haopeng Xie and Cathy Mengying Fang and Tianjian Li and Shayne Longpre and Hongxiang Gu and Maximillian Chen and Tianmin Shu},
      year={2026},
      eprint={2605.20087},
      archivePrefix={arXiv},
      primaryClass={cs.CL},
      url={https://arxiv.org/abs/2605.20087}, 
}

@article{mehri2026measuring,
  title={Measuring and mitigating the distributional gap between real and simulated user behaviors},
  author={Mehri, Shuhaib and Laban, Philippe and Shashidhar, Sumuk and Abdulhai, Marwa and Levine, Sergey and Galley, Michel and Hakkani-T{\"u}r, Dilek},
  journal={arXiv preprint arXiv:2605.07847},
  year={2026}
}

@misc{gemmateam2025gemma3technicalreport,
      title={Gemma 3 Technical Report}, 
      author={Gemma Team and Aishwarya Kamath and Johan Ferret and Shreya Pathak and Nino Vieillard and Ramona Merhej and Sarah Perrin and Tatiana Matejovicova and Alexandre Ramé and Morgane Rivière and Louis Rouillard and Thomas Mesnard and Geoffrey Cideron and Jean-bastien Grill and Sabela Ramos and Edouard Yvinec and Michelle Casbon and Etienne Pot and Ivo Penchev and Gaël Liu and Francesco Visin and Kathleen Kenealy and Lucas Beyer and Xiaohai Zhai and Anton Tsitsulin and Robert Busa-Fekete and Alex Feng and Noveen Sachdeva and Benjamin Coleman and Yi Gao and Basil Mustafa and Iain Barr and Emilio Parisotto and David Tian and Matan Eyal and Colin Cherry and Jan-Thorsten Peter and Danila Sinopalnikov and Surya Bhupatiraju and Rishabh Agarwal and Mehran Kazemi and Dan Malkin and Ravin Kumar and David Vilar and Idan Brusilovsky and Jiaming Luo and Andreas Steiner and Abe Friesen and Abhanshu Sharma and Abheesht Sharma and Adi Mayrav Gilady and Adrian Goedeckemeyer and Alaa Saade and Alex Feng and Alexander Kolesnikov and Alexei Bendebury and Alvin Abdagic and Amit Vadi and András György and André Susano Pinto and Anil Das and Ankur Bapna and Antoine Miech and Antoine Yang and Antonia Paterson and Ashish Shenoy and Ayan Chakrabarti and Bilal Piot and Bo Wu and Bobak Shahriari and Bryce Petrini and Charlie Chen and Charline Le Lan and Christopher A. Choquette-Choo and CJ Carey and Cormac Brick and Daniel Deutsch and Danielle Eisenbud and Dee Cattle and Derek Cheng and Dimitris Paparas and Divyashree Shivakumar Sreepathihalli and Doug Reid and Dustin Tran and Dustin Zelle and Eric Noland and Erwin Huizenga and Eugene Kharitonov and Frederick Liu and Gagik Amirkhanyan and Glenn Cameron and Hadi Hashemi and Hanna Klimczak-Plucińska and Harman Singh and Harsh Mehta and Harshal Tushar Lehri and Hussein Hazimeh and Ian Ballantyne and Idan Szpektor and Ivan Nardini and Jean Pouget-Abadie and Jetha Chan and Joe Stanton and John Wieting and Jonathan Lai and Jordi Orbay and Joseph Fernandez and Josh Newlan and Ju-yeong Ji and Jyotinder Singh and Kat Black and Kathy Yu and Kevin Hui and Kiran Vodrahalli and Klaus Greff and Linhai Qiu and Marcella Valentine and Marina Coelho and Marvin Ritter and Matt Hoffman and Matthew Watson and Mayank Chaturvedi and Michael Moynihan and Min Ma and Nabila Babar and Natasha Noy and Nathan Byrd and Nick Roy and Nikola Momchev and Nilay Chauhan and Noveen Sachdeva and Oskar Bunyan and Pankil Botarda and Paul Caron and Paul Kishan Rubenstein and Phil Culliton and Philipp Schmid and Pier Giuseppe Sessa and Pingmei Xu and Piotr Stanczyk and Pouya Tafti and Rakesh Shivanna and Renjie Wu and Renke Pan and Reza Rokni and Rob Willoughby and Rohith Vallu and Ryan Mullins and Sammy Jerome and Sara Smoot and Sertan Girgin and Shariq Iqbal and Shashir Reddy and Shruti Sheth and Siim Põder and Sijal Bhatnagar and Sindhu Raghuram Panyam and Sivan Eiger and Susan Zhang and Tianqi Liu and Trevor Yacovone and Tyler Liechty and Uday Kalra and Utku Evci and Vedant Misra and Vincent Roseberry and Vlad Feinberg and Vlad Kolesnikov and Woohyun Han and Woosuk Kwon and Xi Chen and Yinlam Chow and Yuvein Zhu and Zichuan Wei and Zoltan Egyed and Victor Cotruta and Minh Giang and Phoebe Kirk and Anand Rao and Kat Black and Nabila Babar and Jessica Lo and Erica Moreira and Luiz Gustavo Martins and Omar Sanseviero and Lucas Gonzalez and Zach Gleicher and Tris Warkentin and Vahab Mirrokni and Evan Senter and Eli Collins and Joelle Barral and Zoubin Ghahramani and Raia Hadsell and Yossi Matias and D. Sculley and Slav Petrov and Noah Fiedel and Noam Shazeer and Oriol Vinyals and Jeff Dean and Demis Hassabis and Koray Kavukcuoglu and Clement Farabet and Elena Buchatskaya and Jean-Baptiste Alayrac and Rohan Anil and Dmitry and Lepikhin and Sebastian Borgeaud and Olivier Bachem and Armand Joulin and Alek Andreev and Cassidy Hardin and Robert Dadashi and Léonard Hussenot},
      year={2025},
      eprint={2503.19786},
      archivePrefix={arXiv},
      primaryClass={cs.CL},
      url={https://arxiv.org/abs/2503.19786}, 
}

@article{mehri2026goal,
  title={Goal alignment in llm-based user simulators for conversational ai},
  author={Mehri, Shuhaib and Yang, Xiaocheng and Kim, Takyoung and Tur, Gokhan and Mehri, Shikib and Hakkani-T{\"u}r, Dilek},
  journal={Transactions of the Association for Computational Linguistics},
  volume={14},
  pages={852--876},
  year={2026},
  publisher={MIT Press}
}

@misc{shao2024deepseekmathpushinglimitsmathematical,
      title={DeepSeekMath: Pushing the Limits of Mathematical Reasoning in Open Language Models}, 
      author={Zhihong Shao and Peiyi Wang and Qihao Zhu and Runxin Xu and Junxiao Song and Xiao Bi and Haowei Zhang and Mingchuan Zhang and Y. K. Li and Y. Wu and Daya Guo},
      year={2024},
      eprint={2402.03300},
      archivePrefix={arXiv},
      primaryClass={cs.CL},
      url={https://arxiv.org/abs/2402.03300}, 
}

@inproceedings{
zhao2026selfdistilled,
title={Self-Distilled Reasoner: On-Policy Self-Distillation for Large Language Models},
author={Siyan Zhao and Zhihui Xie and Mengchen Liu and Jing Huang and Guan Pang and Feiyu Chen and Aditya Grover},
booktitle={Forty-third International Conference on Machine Learning},
year={2026},
url={https://openreview.net/forum?id=Jpxfof0EaS}
}

@misc{sun2026reinforcinghumanbehaviorsimulation,
      title={Reinforcing Human Behavior Simulation via Verbal Feedback}, 
      author={Weiwei Sun and Xuhui Zhou and Jiarui Liu and Weihua Du and Haojia Sun and Yiqing Xie and Qianou Ma and Sihao Chen and Mengting Wan and Longqi Yang and Pei Zhou and Sherry Wu and Sean Welleck and Graham Neubig and Yiming Yang and Maarten Sap},
      year={2026},
      eprint={2605.20506},
      archivePrefix={arXiv},
      primaryClass={cs.LG},
      url={https://arxiv.org/abs/2605.20506}, 
}

@inproceedings{
song2026expanding,
title={Expanding the Capabilities of Reinforcement Learning via Text Feedback},
author={Yuda Song and Lili Chen and Fahim Tajwar and R{\'e}mi Munos and Deepak Pathak and Drew Bagnell and Aarti Singh and Andrea Zanette},
booktitle={The 1st Workshop on Scaling Post-training for LLMs},
year={2026},
url={https://openreview.net/forum?id=om12baYnKM}
}

@InProceedings{pmlr-v15-ross11a,
  title = 	 {A Reduction of Imitation Learning and Structured Prediction to No-Regret Online Learning},
  author = 	 {Ross, Stephane and Gordon, Geoffrey and Bagnell, Drew},
  booktitle = 	 {Proceedings of the Fourteenth International Conference on Artificial Intelligence and Statistics},
  pages = 	 {627--635},
  year = 	 {2011},
  editor = 	 {Gordon, Geoffrey and Dunson, David and Dudík, Miroslav},
  volume = 	 {15},
  series = 	 {Proceedings of Machine Learning Research},
  address = 	 {Fort Lauderdale, FL, USA},
  month = 	 {11--13 Apr},
  publisher =    {PMLR},
  url = 	 {https://proceedings.mlr.press/v15/ross11a.html}
}

@inproceedings{ICLR2024_5be69a58,
 author = {Agarwal, Rishabh and Vieillard, Nino and Zhou, Yongchao and Stanczyk, Piotr and Ramos Garea, Sabela and Geist, Matthieu and Bachem, Olivier},
 booktitle = {International Conference on Learning Representations},
 editor = {B. Kim and Y. Yue and S. Chaudhuri and K. Fragkiadaki and M. Khan and Y. Sun},
 pages = {21246--21263},
 title = {On-Policy Distillation of Language Models: Learning from Self-Generated Mistakes},
 url = {https://proceedings.iclr.cc/paper_files/paper/2024/file/5be69a584901a26c521c2b51e40a4c20-Paper-Conference.pdf},
 volume = {2024},
 year = {2024}
}

@InProceedings{pmlr-v235-xiong24a,
  title = 	 {Iterative Preference Learning from Human Feedback: Bridging Theory and Practice for {RLHF} under {KL}-constraint},
  author =       {Xiong, Wei and Dong, Hanze and Ye, Chenlu and Wang, Ziqi and Zhong, Han and Ji, Heng and Jiang, Nan and Zhang, Tong},
  booktitle = 	 {Proceedings of the 41st International Conference on Machine Learning},
  pages = 	 {54715--54754},
  year = 	 {2024},
  editor = 	 {Salakhutdinov, Ruslan and Kolter, Zico and Heller, Katherine and Weller, Adrian and Oliver, Nuria and Scarlett, Jonathan and Berkenkamp, Felix},
  volume = 	 {235},
  series = 	 {Proceedings of Machine Learning Research},
  month = 	 {21--27 Jul},
  publisher =    {PMLR},
  url = 	 {https://proceedings.mlr.press/v235/xiong24a.html}
}

@inproceedings{
anthis2025position,
title={Position: {LLM} Social Simulations Are a Promising Research Method},
author={Jacy Reese Anthis and Ryan Liu and Sean M Richardson and Austin C. Kozlowski and Bernard Koch and Erik Brynjolfsson and James Evans and Michael S. Bernstein},
booktitle={Forty-second International Conference on Machine Learning Position Paper Track},
year={2025},
url={https://openreview.net/forum?id=cRBg1dtj7o}
}

@inproceedings{
chopra2026cooperativesimulatorsgeneratingrealistic,
title={Beyond Cooperative Simulators: Generating Realistic User Personas for Robust Evaluation of {LLM} Agents},
author={Harshita Chopra and Kshitish Ghate and Aylin Caliskan and Tadayoshi Kohno and Chirag Shah and Natasha Jaques},
booktitle={Second Workshop on Social Simulation with LLMS: Fidelity in Applications},
year={2026},
url={https://openreview.net/forum?id=cJVWKXCws0}
}

@misc{qwen35blog,
    title = {Qwen3.5: Towards Native Multimodal Agents},
    url = {https://qwen.ai/blog?id=qwen3.5},
    author = {Qwen Team},
    month = {February},
    year = {2026}
}

@misc{qwen3embedding,
      title={Qwen3 Embedding: Advancing Text Embedding and Reranking Through Foundation Models}, 
      author={Yanzhao Zhang and Mingxin Li and Dingkun Long and Xin Zhang and Huan Lin and Baosong Yang and Pengjun Xie and An Yang and Dayiheng Liu and Junyang Lin and Fei Huang and Jingren Zhou},
      year={2025},
      eprint={2506.05176},
      archivePrefix={arXiv},
      primaryClass={cs.CL},
      url={https://arxiv.org/abs/2506.05176}, 
}

@inproceedings{10.1145/3209978.3210080,
author = {Zhu, Yaoming and Lu, Sidi and Zheng, Lei and Guo, Jiaxian and Zhang, Weinan and Wang, Jun and Yu, Yong},
title = {Texygen: A Benchmarking Platform for Text Generation Models},
year = {2018},
isbn = {9781450356572},
publisher = {Association for Computing Machinery},
address = {New York, NY, USA},
url = {https://doi.org/10.1145/3209978.3210080},
doi = {10.1145/3209978.3210080},
booktitle = {The 41st International ACM SIGIR Conference on Research \& Development in Information Retrieval},
pages = {1097–1100},
numpages = {4},
location = {Ann Arbor, MI, USA},
series = {SIGIR '18}
}

@misc{xiao2026chameleonslimitinvestigatingpersona,
      title={The Chameleon's Limit: Investigating Persona Collapse and Homogenization in Large Language Models}, 
      author={Yunze Xiao and Vivienne J. Zhang and Chenghao Yang and Ningshan Ma and Weihao Xuan and Jen-tse Huang},
      year={2026},
      eprint={2604.24698},
      archivePrefix={arXiv},
      primaryClass={cs.CL},
      url={https://arxiv.org/abs/2604.24698}, 
}

@inproceedings{lin2004rouge,
  title={Rouge: A package for automatic evaluation of summaries},
  author={Lin, Chin-Yew},
  booktitle={Text summarization branches out},
  pages={74--81},
  year={2004}
}

@inproceedings{
wu2025collabllm,
title={Collab{LLM}: From Passive Responders to Active Collaborators},
author={Shirley Wu and Michel Galley and Baolin Peng and Hao Cheng and Gavin Li and Yao Dou and Weixin Cai and James Zou and Jure Leskovec and Jianfeng Gao},
booktitle={Forty-second International Conference on Machine Learning},
year={2025},
url={https://openreview.net/forum?id=DmH4HHVb3y}
}

@inproceedings{
zhou2024archer,
title={Ar{CH}er: Training Language Model Agents via Hierarchical Multi-Turn {RL}},
author={Yifei Zhou and Andrea Zanette and Jiayi Pan and Sergey Levine and Aviral Kumar},
booktitle={Forty-first International Conference on Machine Learning},
year={2024},
url={https://openreview.net/forum?id=b6rA0kAHT1}
}

@misc{yao2024taubenchbenchmarktoolagentuserinteraction,
      title={$\tau$-bench: A Benchmark for Tool-Agent-User Interaction in Real-World Domains}, 
      author={Shunyu Yao and Noah Shinn and Pedram Razavi and Karthik Narasimhan},
      year={2024},
      eprint={2406.12045},
      archivePrefix={arXiv},
      primaryClass={cs.AI},
      url={https://arxiv.org/abs/2406.12045}, 
}

@inproceedings{ICLR2024_b3075b88,
 author = {Zhou, Xuhui and Zhu, Hao and Mathur, Leena and Zhang, Ruohong and Yu, Haofei and Qi, Zhengyang and Morency, Louis-Philippe and Bisk, Yonatan and Fried, Daniel and Neubig, Graham and Sap, Maarten},
 booktitle = {International Conference on Learning Representations},
 editor = {B. Kim and Y. Yue and S. Chaudhuri and K. Fragkiadaki and M. Khan and Y. Sun},
 pages = {40975--41019},
 title = {SOTOPIA: Interactive Evaluation for Social Intelligence in Language Agents},
 url = {https://proceedings.iclr.cc/paper_files/paper/2024/file/b3075b88e583a0e98d8b24338a613060-Paper-Conference.pdf},
 volume = {2024},
 year = {2024}
}

@inproceedings{liu-etal-2023-one,
    title = "One Cannot Stand for Everyone! Leveraging Multiple User Simulators to train Task-oriented Dialogue Systems",
    author = "Liu, Yajiao  and
      Jiang, Xin  and
      Yin, Yichun  and
      Wang, Yasheng  and
      Mi, Fei  and
      Liu, Qun  and
      Wan, Xiang  and
      Wang, Benyou",
    editor = "Rogers, Anna  and
      Boyd-Graber, Jordan  and
      Okazaki, Naoaki",
    booktitle = "Proceedings of the 61st Annual Meeting of the Association for Computational Linguistics (Volume 1: Long Papers)",
    month = jul,
    year = "2023",
    address = "Toronto, Canada",
    publisher = "Association for Computational Linguistics",
    url = "https://aclanthology.org/2023.acl-long.1/",
    doi = "10.18653/v1/2023.acl-long.1",
    pages = "1--21"
}

@inproceedings{10.1145/3586183.3606763,
author = {Park, Joon Sung and O'Brien, Joseph and Cai, Carrie Jun and Morris, Meredith Ringel and Liang, Percy and Bernstein, Michael S.},
title = {Generative Agents: Interactive Simulacra of Human Behavior},
year = {2023},
isbn = {9798400701320},
publisher = {Association for Computing Machinery},
address = {New York, NY, USA},
url = {https://doi.org/10.1145/3586183.3606763},
doi = {10.1145/3586183.3606763},
booktitle = {Proceedings of the 36th Annual ACM Symposium on User Interface Software and Technology},
articleno = {2},
numpages = {22},
location = {San Francisco, CA, USA},
series = {UIST '23}
}

@misc{wang2026learningusersimulatorsturing,
      title={Learning User Simulators with Turing Rewards}, 
      author={Yingshan Susan Wang and Cedegao E. Zhang and Linlu Qiu and Zexue He and Pengyuan Li and Alex Pentland and Roger P. Levy and Yoon Kim},
      year={2026},
      eprint={2606.19336},
      archivePrefix={arXiv},
      primaryClass={cs.CL},
      url={https://arxiv.org/abs/2606.19336}, 
}

@inproceedings{wang-etal-2025-know,
    title = "Know You First and Be You Better: Modeling Human-Like User Simulators via Implicit Profiles",
    author = "Wang, Kuang  and
      Li, Xianfei  and
      Yang, Shenghao  and
      Zhou, Li  and
      Jiang, Feng  and
      Li, Haizhou",
    editor = "Che, Wanxiang  and
      Nabende, Joyce  and
      Shutova, Ekaterina  and
      Pilehvar, Mohammad Taher",
    booktitle = "Proceedings of the 63rd Annual Meeting of the Association for Computational Linguistics (Volume 1: Long Papers)",
    month = jul,
    year = "2025",
    address = "Vienna, Austria",
    publisher = "Association for Computational Linguistics",
    url = "https://aclanthology.org/2025.acl-long.1025/",
    doi = "10.18653/v1/2025.acl-long.1025",
    pages = "21082--21107",
    ISBN = "979-8-89176-251-0"
}

@inproceedings{li-etal-2016-persona,
    title = "A Persona-Based Neural Conversation Model",
    author = "Li, Jiwei  and
      Galley, Michel  and
      Brockett, Chris  and
      Spithourakis, Georgios  and
      Gao, Jianfeng  and
      Dolan, Bill",
    editor = "Erk, Katrin  and
      Smith, Noah A.",
    booktitle = "Proceedings of the 54th Annual Meeting of the Association for Computational Linguistics (Volume 1: Long Papers)",
    month = aug,
    year = "2016",
    address = "Berlin, Germany",
    publisher = "Association for Computational Linguistics",
    url = "https://aclanthology.org/P16-1094/",
    doi = "10.18653/v1/P16-1094",
    pages = "994--1003"
}

@inproceedings{schatzmann-etal-2007-agenda,
    title = "Agenda-Based User Simulation for Bootstrapping a {POMDP} Dialogue System",
    author = "Schatzmann, Jost  and
      Thomson, Blaise  and
      Weilhammer, Karl  and
      Ye, Hui  and
      Young, Steve",
    editor = "Sidner, Candace  and
      Schultz, Tanja  and
      Stone, Matthew  and
      Zhai, ChengXiang",
    booktitle = "Human Language Technologies 2007: The Conference of the North {A}merican Chapter of the Association for Computational Linguistics; Companion Volume, Short Papers",
    month = apr,
    year = "2007",
    address = "Rochester, New York",
    publisher = "Association for Computational Linguistics",
    url = "https://aclanthology.org/N07-2038/",
    pages = "149--152"
}

@article{Argyle_Busby_Fulda_Gubler_Rytting_Wingate_2023, title={Out of One, Many: Using Language Models to Simulate Human Samples}, volume={31}, DOI={10.1017/pan.2023.2}, number={3}, journal={Political Analysis}, author={Argyle, Lisa P. and Busby, Ethan C. and Fulda, Nancy and Gubler, Joshua R. and Rytting, Christopher and Wingate, David}, year={2023}, pages={337–351}}

@ARTICLE{2024arXiv241110109P,
       author = {{Park}, Joon Sung and {Zou}, Carolyn Q. and {Kamphorst}, Jonne and {Egan}, Niles and {Shaw}, Aaron and {Mako Hill}, Benjamin and {Cai}, Carrie and {Ringel Morris}, Meredith and {Liang}, Percy and {Willer}, Robb and {Bernstein}, Michael S.},
        title = "{LLM Agents Grounded in Self-Reports Enable General-Purpose Simulation of Individuals}",
      journal = {arXiv e-prints},
         year = 2024,
        month = nov,
          eid = {arXiv:2411.10109},
        pages = {arXiv:2411.10109},
          doi = {10.48550/arXiv.2411.10109},
archivePrefix = {arXiv},
       eprint = {2411.10109},
 primaryClass = {cs.AI},
       adsurl = {https://ui.adsabs.harvard.edu/abs/2024arXiv241110109P}
}

@misc{chen2025personavectorsmonitoringcontrolling,
      title={Persona Vectors: Monitoring and Controlling Character Traits in Language Models}, 
      author={Runjin Chen and Andy Arditi and Henry Sleight and Owain Evans and Jack Lindsey},
      year={2025},
      eprint={2507.21509},
      archivePrefix={arXiv},
      primaryClass={cs.CL},
      url={https://arxiv.org/abs/2507.21509}, 
}

@inproceedings{10.1145/3701716.3715463,
author = {Ning, Lin and Liu, Luyang and Wu, Jiaxing and Wu, Neo and Berlowitz, Devora and Prakash, Sushant and Green, Bradley and O'Banion, Shawn and Xie, Jun},
title = {User-LLM: Efficient LLM Contextualization with User Embeddings},
year = {2025},
isbn = {9798400713316},
publisher = {Association for Computing Machinery},
address = {New York, NY, USA},
url = {https://doi.org/10.1145/3701716.3715463},
doi = {10.1145/3701716.3715463},
booktitle = {Companion Proceedings of the ACM on Web Conference 2025},
pages = {1219–1223},
numpages = {5},
location = {Sydney NSW, Australia},
series = {WWW '25}
}

@inproceedings{NEURIPS2025_5fc150b7,
 author = {Li, Ang and Chen, Haozhe and Namkoong, Hongseok and Peng, Tianyi},
 booktitle = {Advances in Neural Information Processing Systems},
 doi = {10.52202/085713-2226},
 editor = {D. Belgrave and C. Zhang and H. Lin and R. Pascanu and P. Koniusz and M. Ghassemi and N. Chen},
 pages = {},
 publisher = {Curran Associates, Inc.},
 title = {LLM Generated Persona is a Promise with a Catch},
 url = {https://proceedings.neurips.cc/paper_files/paper/2025/file/5fc150b75e8b509735564de16bc3527b-Paper-Position_Paper_Track.pdf},
 volume = {38, Main Conference},
 year = {2025}
}

@misc{liu2026individualpersonasaligningsynthetic,
      title={Beyond Individual Personas: Aligning Synthetic Dialogue to Population-Level Behavior Distributions}, 
      author={Xinyi Liu and Rinat Khaziev and Hooshang Nayyeri and Emine Yilmaz and Charith Peris and Hari Thadakamalla},
      year={2026},
      eprint={2606.07893},
      archivePrefix={arXiv},
      primaryClass={cs.CL},
      url={https://arxiv.org/abs/2606.07893}, 
}

@inproceedings{
lin2026the,
title={The Illusion of Intervention: Your {LLM}-Simulated Experiment is an Observational Study},
author={Victoria Lin and Taedong Yun and Maja Mataric and John Canny and Arthur Gretton and Alexander Nicholas D'Amour},
booktitle={ICML 2026 Statistical Frameworks for Uncertainty in Agentic Systems},
year={2026},
url={https://openreview.net/forum?id=7EZrgYo07P}
}

@inproceedings{
shi2026experiential,
title={Experiential Reinforcement Learning},
author={Taiwei Shi and Sihao Chen and Bowen Jiang and Linxin Song and Longqi Yang and Jieyu Zhao},
booktitle={ICLR 2026 Workshop on Lifelong Agents: Learning, Aligning, Evolving},
year={2026},
url={https://openreview.net/forum?id=OSYyL6hqdP}
}

@inproceedings{
hubotter2026reinforcement,
title={Reinforcement Learning via Self-Distillation},
author={Jonas H{\"u}botter and Frederike L{\"u}beck and Lejs Deen Behric and Anton Baumann and Marco Bagatella and Daniel Marta and Ido Hakimi and Idan Shenfeld and Thomas Kleine Buening and Carlos Guestrin and Andreas Krause},
booktitle={Forty-third International Conference on Machine Learning},
year={2026},
url={https://openreview.net/forum?id=QkfkxyRizZ}
}

@inproceedings{
mendes2026making,
title={Making Expert Reasoning Learnable with Self-Distillation},
author={Ethan Mendes and Jungsoo Park and Alan Ritter},
booktitle={Forty-third International Conference on Machine Learning},
year={2026},
url={https://openreview.net/forum?id=JG6f02X29a}
}

@misc{buening2026aligninglanguagemodelsuser,
      title={Aligning Language Models from User Interactions}, 
      author={Thomas Kleine Buening and Jonas Hübotter and Barna Pásztor and Idan Shenfeld and Giorgia Ramponi and Andreas Krause},
      year={2026},
      eprint={2603.12273},
      archivePrefix={arXiv},
      primaryClass={cs.CL},
      url={https://arxiv.org/abs/2603.12273}, 
}

@misc{wang2026skillsdskillconditionedselfdistillationmultiturn,
      title={Skill-SD: Skill-Conditioned Self-Distillation for Multi-turn LLM Agents}, 
      author={Hao Wang and Guozhi Wang and Han Xiao and Yufeng Zhou and Yue Pan and Jichao Wang and Ke Xu and Yafei Wen and Xiaohu Ruan and Xiaoxin Chen and Honggang Qi},
      year={2026},
      eprint={2604.10674},
      archivePrefix={arXiv},
      primaryClass={cs.LG},
      url={https://arxiv.org/abs/2604.10674}, 
}

@misc{wang2026hindsighthintdistillationscaffolded,
      title={Hindsight Hint Distillation: Scaffolded Reasoning for SWE Agents from CoT-free Answers}, 
      author={Shengjie Wang and Guanghe Li and Zonghan Yang and Yang Gao},
      year={2026},
      eprint={2605.11556},
      archivePrefix={arXiv},
      primaryClass={cs.AI},
      url={https://arxiv.org/abs/2605.11556}, 
}

@misc{yeo2026hintsdtargetedhindsightselfdistillation,
      title={HINT-SD: Targeted Hindsight Self-Distillation for Long-Horizon Agents}, 
      author={Woongyeong Yeo and Yumin Choi and Taekyung Ki and Sung Ju Hwang},
      year={2026},
      eprint={2605.17873},
      archivePrefix={arXiv},
      primaryClass={cs.LG},
      url={https://arxiv.org/abs/2605.17873}, 
}

@misc{li2026distillselectivehindsightdistillation,
      title={What and When to Distill: Selective Hindsight Distillation for Multi-Turn Agents}, 
      author={Xiaozhe Li and Tianyi Lyu and Yang Li and Yichuan Ma and Peiji Li and Linyang Li and Qipeng Guo and Dahua Lin and Kai Chen},
      year={2026},
      eprint={2605.19447},
      archivePrefix={arXiv},
      primaryClass={cs.AI},
      url={https://arxiv.org/abs/2605.19447}, 
}

@misc{liu2026herohindsightenhancedreflectionenvironment,
      title={HERO: Hindsight-Enhanced Reflection from Environment Observations for Agentic Self-Distillation}, 
      author={Haoran Liu and Yuwei Zhang and Xiyao Li and Bohan Lyu and Jingbo Shang},
      year={2026},
      eprint={2606.11559},
      archivePrefix={arXiv},
      primaryClass={cs.AI},
      url={https://arxiv.org/abs/2606.11559}, 
}

@misc{mehri2026multisessioncollablearninguserpreferences,
      title={MultiSessionCollab: Learning User Preferences with Memory to Improve Long-Term Collaboration}, 
      author={Shuhaib Mehri and Priyanka Kargupta and Tal August and Dilek Hakkani-Tür},
      year={2026},
      eprint={2601.02702},
      archivePrefix={arXiv},
      primaryClass={cs.AI},
      url={https://arxiv.org/abs/2601.02702}, 
}
\bibliographystyle{iclr2027_conference}

\clearpage
\newpage

\appendix

\section{Hindsight Distillation Algorithm}
\label{appendix:algm}

Algorithm~\ref{alg:hindsight-distillation} presents our hindsight distillation algorithm. This algorithm distills hindsight feedback-conditioned improvements into a representation generator model, enabling it to generate improved representations directly from an input conversation.

\begin{algorithm}[h]
    \caption{Hindsight Distillation}
    \label{alg:hindsight-distillation}
    \begin{algorithmic}[1]
        \STATE \textbf{input:} representation generator $G_\theta$;
        conversation dataset $\mathcal{D}$; ensemble of simulation environments $\mathcal{E}=\{(U_k,A_k,J_k)\}_{k=1}^{K}$; refinement rounds $I$
        \REPEAT
            \STATE sample a conversation $\mathcal{C}\sim\mathcal{D}$.
            \STATE sample an initial user representation $r^{(0)}\sim G_\theta(\cdot\mid\mathcal{C})$.
            \FOR{$i=0,\ldots,I-1$}
                \FORALL{$e_k=(U_k,A_k,J_k)\in\mathcal{E}$}
                    \STATE roll out
                    $\tilde{\mathcal{C}}_k^{(i)}
                    \sim\operatorname{rollout}
                    (U_k(\cdot\mid r^{(i)}),A_k)$.
                    \STATE generate hindsight feedback
                    $f_k^{(i)}
                    \sim J_k(\cdot\mid
                    \mathcal{C},\tilde{\mathcal{C}}_k^{(i)})$.
                \ENDFOR
                \STATE collect feedback
                $f^{(i)}=(f_1^{(i)},\ldots,f_K^{(i)})$.
                \STATE sample a refined user representation
                $r^{(i+1)}
                \sim G_\theta(\cdot\mid
                \mathcal{C},r^{(i)},f^{(i)})$.
            \ENDFOR
            \STATE update $\theta$ with gradient descent on $\mathcal{L}(\theta)$ to predict $r^{(I)}$ from $\mathcal{C}$ alone.
        \UNTIL{converged}
    \end{algorithmic}
\end{algorithm}

\section{Human Validation of LLM Judge}
\label{appendix:judge-validation}

In this section, we validate whether GPT-5.4 mini's fidelity scores align with human judgements. We construct a human-labeled validation set of 100 original conversations from our evaluation set, each paired with two simulations generated using representations from Qwen3.5-4B and \texttt{UserRepGen-4B}, with GPT-5.4 mini as both user simulator and assistant. We select only pairs for which the LLM judge assigns unequal fidelity scores. For each pair, presented in randomized order and without its source or scores, we label the simulation whose user behavior is more faithful to the original user. We then measure how often the judge's ranking agrees with these labels.

The judge's ranking agrees with our labels on 88\% of pairs. As shown in Table~\ref{tab:judge-human-agreement}, agreement increases with the absolute difference between the two fidelity scores, from 80.5\% when the gap is one point to 97.4\% when the gap is at least three points.

\begin{table}[h]
    \centering
    \caption{Agreement between our manually labeled preferences and the LLM judge. The score gap is the absolute difference between the two simulated conversations' overall fidelity scores.}
    \label{tab:judge-human-agreement}
    \small
    \begin{tabular}{lcc}
        \toprule
        \textbf{Score gap} & \textbf{Agreement} & \textbf{Rate} \\
        \midrule
        1 & 33 / 41 & 80.5\% \\
        2 & 17 / 20 & 85.0\% \\
        $\geq 3$ & 38 / 39 & 97.4\% \\
        \midrule
        Overall & 88 / 100 & 88.0\% \\
        \bottomrule
    \end{tabular}
\end{table}

These results support the LLM judge as a measure of simulation fidelity. Moreover, our evaluations for population-level diversity and distinguishing users across conversations in Sections \ref{sec:population-level} and \ref{sec:user-reid} provide complementary evidence, by demonstrating that representations optimized for simulation fidelity improve performance on downstream tasks that do not rely on the fidelity judge.

\section{Prompts}
\label{appendix:prompts}

\begin{tcolorbox}[
    colback=gray!5,
    colframe=gray!75!black,
    title=User Representation Generation Prompt,
    fonttitle=\bfseries,
    left=2mm,
    right=2mm,
    top=1mm,
    bottom=1mm,
    arc=2mm,
    boxrule=0.5pt,
    breakable
]
\begin{lstlisting}[
    basicstyle=\scriptsize\ttfamily,
    breaklines=true,
]
You are an expert at analyzing user behavior in conversations between a user and an AI assistant. Given the conversation below, your task is to produce a representation of the user.

Treat the conversation as evidence about the user, and use it to infer how they generally behave and how they would typically act in other conversations. Focus on stable, general tendencies such as: how they phrase and specify requests, how they decompose goals across turns, how they respond to and give feedback to the assistant, how and when they provide context, their communication style, the dialog acts they use, and their degree and depth of engagement. Keep this description general: do not include topic-specific details or content from the conversation, do not quote or closely paraphrase full user messages, and do not refer to specific moments in this conversation (e.g. "here", "in this conversation", "when they asked about X"). Short distinctive words or phrases may be quoted sparingly when they are useful examples of the user's communication style. Describe their patterns in your own general terms instead.

Separately, you will also describe the task the user is trying to accomplish. Here you may include topic-specific details about what the user is doing in the conversation.

Conversation:
{conversation_history}

Output a single valid JSON object of the form {{"task": "...", "user_behavior": "..."}}. Do not wrap the output in markdown or add any other text.
\end{lstlisting}
\end{tcolorbox}

\begin{tcolorbox}[
    colback=gray!5,
    colframe=gray!75!black,
    title=Improving User Representation Prompt,
    fonttitle=\bfseries,
    left=2mm,
    right=2mm,
    top=1mm,
    bottom=1mm,
    arc=2mm,
    boxrule=0.5pt,
    breakable
]
\begin{lstlisting}[
    basicstyle=\scriptsize\ttfamily,
    breaklines=true,
]
You are an expert at analyzing user behavior in conversations between a user and an AI assistant. You have a representation of the user based on the conversation below. We used that representation to simulate a user, and then collected feedback from a judge. Your task is to incorporate that feedback and generate an improved user representation.

The user representation consists of a task and user_behavior field. The task field describes what the user is trying to accomplish, and can include topic-specific details about what the user is doing in the conversation. The user_behavior field describes how they generally behave and how they would typically act in conversations. It focuses on stable, general tendencies such as: how they phrase and specify requests, how they decompose goals across turns, how they respond to and give feedback to the assistant, how and when they provide context, their communication style, the dialog acts they use, and their degree and depth of engagement. This description should be general: it should not include topic-specific details or content from the conversation, it should not quote or closely paraphrase full user messages, and it should not refer to specific moments in this conversation (e.g. "here", "in this conversation", "when they asked about X"). Short distinctive words or phrases may be quoted sparingly when they are useful examples of the user's communication style. It should describe their patterns in general terms.

Additional guidelines for improving the representation:
- Optimize for behavioral simulation, not efficient task completion: preserve multi-turn repetition, incremental follow-ups, verbose constraints, and inefficient interaction patterns when they appear in the real user.
- Make user_behavior operational: specify what the simulator should do and should not do, including negative constraints from judge feedback (e.g. do not terminate early, combine turns, over-polish, change formatting style, or solve the task directly).
- Be explicit about role distinctions: if the real user asks the assistant/agent to produce commands, JSON, code, tool calls, or formatted output, say whether the user writes instructions to the assistant or directly provides the final artifact.
- When feedback says the simulation is too short, too generic, or collapsed multiple turns, specify the desired utterance form and interaction arc: approximate turn count, near-duplicate vs incremental turns, what each turn should contain, and what not to compress.
- For prompt-author or agent-control users, state if user turns should be long prompt blocks with role instructions, constraints, allowed actions, response schemas, evaluation criteria, and goals; also state not to replace those blocks with a bare command, JSON object, or final answer.

Conversation:
{conversation_history}

Current User Representation:
task: {task}
user_behavior: {user_behavior}

Judge Feedback:
{feedback}

Generate an updated representation that incorporates the judge feedback. This updated representation should follow the guidelines above, and should be in the same format as the current representation.
Output only valid JSON with exactly this shape:
{{"task": "<improved task>", "user_behavior": "<improved behavior representation>"}}
\end{lstlisting}
\end{tcolorbox}

\begin{tcolorbox}[
    colback=gray!5,
    colframe=gray!75!black,
    title=User Simulator System Prompt,
    fonttitle=\bfseries,
    left=2mm,
    right=2mm,
    top=1mm,
    bottom=1mm,
    arc=2mm,
    boxrule=0.5pt,
    breakable
]
\begin{lstlisting}[
    basicstyle=\scriptsize\ttfamily,
    breaklines=true,
]
You are a user simulator. You will be provided with a user representation and your task is to role-play as that user in a conversation with an AI assistant.

# User Representation
{user_representation}

# Instructions
- Stay consistent with the information provided in the user representation
- Never describe the user or your own behavior, never break character, and never mention that you are simulating, role-playing, or following a description.
- When the task requires concrete material for the assistant to work on (e.g. a paragraph to analyze, a passage to translate, code to debug, an essay to edit, data to summarize), invent a plausible, self-contained example yourself and include it inline in your message. Never ask the assistant to work on something you have not actually provided.
- When this user's goal has been met, or when they would naturally end the chat, respond with exactly \"{termination_signal}\" and nothing else.\end{lstlisting}
\end{tcolorbox}

\begin{tcolorbox}[
    colback=gray!5,
    colframe=gray!75!black,
    title=Assistant System Prompt,
    fonttitle=\bfseries,
    left=2mm,
    right=2mm,
    top=1mm,
    bottom=1mm,
    arc=2mm,
    boxrule=0.5pt,
    breakable
]
\begin{lstlisting}[
    basicstyle=\scriptsize\ttfamily,
    breaklines=true,
]
You are a helpful AI assistant. Respond to the user's messages accurately and helpfully.
\end{lstlisting}
\end{tcolorbox}

\begin{tcolorbox}[
    colback=gray!5,
    colframe=gray!75!black,
    title=Judge Prompt - Full Conversation,
    fonttitle=\bfseries,
    left=2mm,
    right=2mm,
    top=1mm,
    bottom=1mm,
    arc=2mm,
    boxrule=0.5pt,
    breakable
]
\begin{lstlisting}[
    basicstyle=\scriptsize\ttfamily,
    breaklines=true,
]
You are an expert analyzing user behavior in human-AI conversations. Your task is to evaluate how well a simulated user reproduces the behavior of a real user.

You are given the full real and simulated conversations, including assistant turns. Use the assistant turns only as context for interpreting the user's behavior; score only how faithfully the SIMULATED user reproduces the REAL user's behavior. Do not score assistant quality.

Evaluate how well the simulated user reproduces the real user along the following dimensions:

- Communication style:
    - Do they use similar punctuation, capitalization, spelling/grammar, typos, slang, emoji, contractions, and other surface-level writing habits? Compare concrete surface artifacts, not just the general vibe.
    - Do they have similar response lengths? Are they terse or verbose in the same way?
    - Do they use similar formatting, such as paragraphs, markdown, bullets, or line breaks?
    - Do they have similar tone/formality, politeness, greetings/thanks, directness, warmth, and social conventions?
- Engagement (The REAL user took {real_user_turns} turn(s); the SIMULATED user took {sim_user_turns} turn(s)):
    - Do they spend a similar number of turns on the same interaction arc or subtask? Do they persist or move on quickly in a similar way?
    - Do they continue, stop, or terminate at a similar point?
- Requests:
    - Do they pursue a similar number of distinct asks or subtasks? Ignore the exact domain/topic, but compare the interaction structure.
    - Do they ask for things in a similar way: direct commands vs questions, broad asks vs narrow asks, explicit constraints vs implicit cues?
    - Do they front-load requirements or reveal them gradually across turns in a similar way?
    - Do they keep separate real-user requests separate, rather than combining them too early?
    - Are their requests similarly open-ended, verification-oriented, correction-oriented, or solution-oriented?
- Responses:
    - Given the assistant turns shown in each transcript, do the users respond to assistant outputs in similar ways?
    - Do they accept, reject, correct, validate, express dissatisfaction, refine constraints, or redirect the conversation similarly?
    - Is their feedback similarly explicit or implicit, terse or explanatory, polite or blunt?
- Context:
    - Do they provide a similar amount and type of context: examples, data, code, prior attempts, constraints, personal background?
    - Is context front-loaded or revealed gradually in a similar way?

# REAL conversation
{real}

# SIMULATED conversation
{simulated}

# Additional instructions
- Focus ONLY on the USER behavior, and judge HOW the user behaves rather than WHAT they are working on: the specific domain, entities, technologies, or subject matter are not important. Do not reward or penalize the simulated user for working on the same or a different topic than the real user.
- Use assistant turns only to understand what the user is responding to. Do not penalize the simulated user because the assistant gave different content, unless the simulated user's behavioral response pattern is also different.
- Account for the fact that the simulated user may need to act differently because the simulated assistant behaved differently. A faithful user may ask a different follow-up, skip a correction, or introduce a clarification at a different point if the assistant's answer created a different opportunity; judge whether the underlying behavioral tendency is similar.
- Pay attention to whether the simulated user reproduces the full arc of the real user's user turns. For example, if the real user works on multiple tasks across the conversation but the simulated user works on only one, there is a behavioral mismatch.
- Penalize generic simulator behavior such as asking "How can I help?", ending immediately, over-polishing terse users, using markdown when the real user does not, or combining multiple real-user follow-ups into one turn.

For each dimension, write a one-sentence rationale, and then assign a score from 1 (poorly reproduces the real user's behavior) to 10 (highly reproduces the real user's behavior). Finally, give a single 1-10 overall score that measures the simulator's overall behavioral similarity.

Output a valid JSON object with exactly this shape and nothing else (no markdown):
{{"dimensions": {{"communication_style": {{"rationale": "<one sentence>", "score": <1-10>}}, "engagement": {{"rationale": "<one sentence>", "score": <1-10>}}, "requests": {{"rationale": "<one sentence>", "score": <1-10>}}, "responses": {{"rationale": "<one sentence>", "score": <1-10>}}, "context": {{"rationale": "<one sentence>", "score": <1-10>}}}}, "overall": <1-10>}}
\end{lstlisting}
\end{tcolorbox}

\begin{tcolorbox}[
    colback=gray!5,
    colframe=gray!75!black,
    title=Judge Prompt - User Utterances Only,
    fonttitle=\bfseries,
    left=2mm,
    right=2mm,
    top=1mm,
    bottom=1mm,
    arc=2mm,
    boxrule=0.5pt,
    breakable
]
\begin{lstlisting}[
    basicstyle=\scriptsize\ttfamily,
    breaklines=true,
]
You are an expert analyzing user behavior in human-AI conversations. Your task is to evaluate how well a simulated user reproduces the behavior of a real user.

You are given the user utterances from a real and simulated conversation between a user and an AI assistant. Evaluate how well the simulated user reproduces the behavior of the real user along the following dimensions:

- Communication style:
    - Do they use similar punctuation, capitalization, spelling/grammar, typos, slang, emoji, contractions, and other surface-level writing habits? Compare concrete surface artifacts, not just the general vibe.
    - Do they have similar response lengths? Are they terse or verbose in the same way?
    - Do they use similar formatting, such as paragraphs, markdown, bullets, or line breaks?
    - Do they have similar tone/formality, politeness, greetings/thanks, directness, warmth, and social conventions?
- Engagement (The REAL user took {real_user_turns} turn(s); the SIMULATED user took {sim_user_turns} turn(s)):
    - Do they spend a similar number of turns on the same interaction arc or subtask? Do they persist or move on quickly in a similar way?
    - Do they continue, stop, or terminate at a similar point?
- Requests:
    - Do they pursue a similar number of distinct asks or subtasks? Ignore the exact domain/topic, but compare the interaction structure.
    - Do they ask for things in a similar way: direct commands vs questions, broad asks vs narrow asks, explicit constraints vs implicit cues?
    - Do they front-load requirements or reveal them gradually across turns in a similar way?
    - Do they keep separate real-user requests separate, rather than combining them too early?
    - Are their requests similarly open-ended, verification-oriented, correction-oriented, or solution-oriented?
- Responses:
    - Based only on the visible user follow-up turns, do they appear to address/engage with prior assistant outputs in a similar way?
    - Do they accept, reject, correct, validate, express dissatisfaction, refine constraints, or redirect the conversation similarly?
    - Is their feedback similarly explicit or implicit, terse or explanatory, polite or blunt?
- Context:
    - Do they provide a similar amount and type of context: examples, data, code, prior attempts, constraints, personal background?
    - Is context front-loaded or revealed gradually in a similar way?

# REAL user utterances
{real}

# SIMULATED user utterances
{simulated}

# Additional instructions
- Focus ONLY on the USER behavior, and judge HOW the user behaves rather than WHAT they are working on: the specific domain, entities, technologies, or subject matter are not important. Do not reward or penalize the simulated user for working on the same or a different topic than the real user.
- Pay attention to whether the simulated user reproduces the full arc of the real user's user turns. For example, if the real user works on multiple tasks across the conversation but the simulated user works on only one, there is a behavioral mismatch.
- Penalize generic simulator behavior such as asking "How can I help?", ending immediately, over-polishing terse users, using markdown when the real user does not, or combining multiple real-user follow-ups into one turn.
- Do not infer hidden assistant content. If a response pattern is not visible from the user utterances themselves, do not give credit for it.

For each dimension, write a one-sentence rationale, and then assign a score from 1 (poorly reproduces the real user's behavior) to 10 (highly reproduces the real user's behavior). Finally, give a single 1-10 overall score that measures the simulator's overall behavioral similarity.

Output a valid JSON object with exactly this shape and nothing else (no markdown):
{{"dimensions": {{"communication_style": {{"rationale": "<one sentence>", "score": <1-10>}}, "engagement": {{"rationale": "<one sentence>", "score": <1-10>}}, "requests": {{"rationale": "<one sentence>", "score": <1-10>}}, "responses": {{"rationale": "<one sentence>", "score": <1-10>}}, "context": {{"rationale": "<one sentence>", "score": <1-10>}}}}, "overall": <1-10>}}
\end{lstlisting}
\end{tcolorbox}

\section{Training Hyperparameters}
\label{appendix:hyperparams}

\begin{table}[H]
\centering
\caption{Training hyperparameters for GRPO, OPSD, SFT, and OSFT.}
\label{tab:training_config}
\small
\begin{tabular}{lcccc}
\toprule
\textbf{Hyperparameter} & \textbf{GRPO} & \textbf{OPSD} & \textbf{SFT} & \textbf{OSFT} \\
\midrule
Learning rate
    & $5 \times 10^{-6}$
    & $5 \times 10^{-6}$
    & $5 \times 10^{-6}$
    & $5 \times 10^{-6}$ \\
Effective batch size
    & 32 & 32 & 32 & 32 \\
Training steps
    & 100 & 100 & 100 & 100 \\
\midrule
LoRA rank ($r$)
    & 32 & 64 & 64 & 64 \\
LoRA alpha ($\alpha$)
    & 64 & 128 & 128 & 128 \\
\midrule
Maximum prompt length
    & 8,192 & 8,192 & 8,192 & 8,192 \\
Maximum completion length
    & 1,024 & 1,024 & 1,024 & 1,024 \\
Maximum sequence length
    & 9,216 & 9,216 & 9,216 & 9,216 \\
\midrule
Generations per prompt
    & 8 & 1 & -- & 1 \\
Sampling temperature
    & 0.7 & 1.1 & -- & 1.1 \\
Top-$p$
    & 0.8 & 0.95 & -- & 0.95 \\
Top-$k$
    & 20 & 20 & -- & 20 \\

\bottomrule
\end{tabular}
\end{table}

\section{Further Analysis on How Training Changes User Representations}
\label{appendix:rep-analysis}

\subsection{Pairwise Analysis with GPT-5.4 Mini}

In Section~\ref{sec:analyzing-user-reps}, we qualitatively analyze how training changes user representations by manually comparing representations generated by the base and trained models. We use the resulting observations to guide a larger-scale, systematic analysis over the 700 conversations in our evaluation set. For each conversation, we present GPT-5.4 mini with the representations generated by Qwen3.5-4B and \texttt{UserRepGen-4B} in randomized order, and ask it to answer 11 questions about how they differ:

\noindent\textit{Task descriptions.}
\begin{enumerate}[leftmargin=18pt,topsep=0pt,itemsep=0pt]
    \item Which task description better captures constraints introduced in later turns?
    \item Which task description better captures shifts in direction or goal?
    \item Which task description better captures multiple distinct objectives or subtasks?
    \item Which task description has less repetition?
\end{enumerate}

\noindent\textit{Behavior descriptions.}
\begin{enumerate}[leftmargin=18pt,topsep=0pt,itemsep=0pt]
    \setcounter{enumi}{4}
    \item Which behavior description is more specific and actionable?
    \item Which behavior description uses more useful negative and contrastive instructions?
    \item Which behavior description better captures the user's response policy?
    \item Which behavior description better captures the user's communication style?
    \item Which behavior description better captures the user's engagement patterns, including how long they persist, when they move on, and when they end the conversation?
    \item Which behavior description better captures how the user makes requests, including how many requests or subtasks they pursue, how directly they ask, and whether they provide requirements upfront or gradually?
    \item Which behavior description better captures how the user provides context, including the amount and type of context and whether it is provided upfront or gradually?
\end{enumerate}

For each question, GPT-5.4 mini must select one of the two representations. We use the following prompt:

\newpage
\begin{tcolorbox}[
    colback=gray!5,
    colframe=gray!75!black,
    title=How Training Changes User Representations - Judge Prompt,
    fonttitle=\bfseries,
    left=2mm,
    right=2mm,
    top=1mm,
    bottom=1mm,
    arc=2mm,
    boxrule=0.5pt,
    breakable
]
\begin{lstlisting}[
    basicstyle=\scriptsize\ttfamily,
    breaklines=true,
]
You are an expert in analyzing user representations. A user representation is a natural-language description generated from a conversation and used to condition an LLM-based user simulator. Its purpose is to give the simulator the information needed to reproduce the user behavior observed in the original conversation. User representations are made up of two components: (1) a task description that describes what the user is trying to accomplish, and (2) a behavior description that describes how the user behaves.

Your task is to help me analyze the difference between the two user representations. You will be given two representations generated from the same conversation, presented in randomized order. The order carries no information about which models produced them or which representation is better. Carefully compare the content of the two representations for every criterion and answer the questions below.

# REPRESENTATION 1
task: {representation_1_task}
behavior: {representation_1_behavior}

# REPRESENTATION 2
task: {representation_2_task}
behavior: {representation_2_behavior}

# For each criterion of the following criteria, select either `representation_1` or `representation_2`:

## Task Description
1. Which representation better captures constraints introduced in later turns?
2. Which representation better captures shifts in direction or goal?
3. Which representation better captures multiple distinct objectives or subtasks?
4. Which representation has less repetition?

## Behavior Description
5. Which behavior description is more specific and actionable? Prefer descriptions of observable patterns that a simulator can reproduce over broad characterizations such as "goal-oriented" or "directive".
6. Which behavior description uses more useful negative and contrastive instructions?
7. Which behavior description better captures the user's response policy?
8. Which behavior description better captures the user's communication style?
9. Which behavior description better captures the user's engagement patterns, including how long they persist, when they move on, and when they end the conversation?
10. Which behavior description better captures how the user makes requests, including how many requests or subtasks they pursue, how directly they ask, and whether they provide requirements upfront or gradually?
11. Which behavior description better captures how the user provides context, including the amount and type of context and whether it is provided upfront or gradually?

# Output format

Output only the following JSON object and nothing else. For each question, provide your selection and one brief sentence explaining the most important difference:

{{
  "q1": {{"selection": "representation_1|representation_2", "reasoning": "<brief sentence>"}}, "q2": {{"selection": "representation_1|representation_2", "reasoning": "<brief sentence>"}}, "q3": {{"selection": "representation_1|representation_2", "reasoning": "<brief sentence>"}}, "q4": {{"selection": "representation_1|representation_2", "reasoning": "<brief sentence>"}}, "q5": {{"selection": "representation_1|representation_2", "reasoning": "<brief sentence>"}}, "q6": {{"selection": "representation_1|representation_2", "reasoning": "<brief sentence>"}}, "q7": {{"selection": "representation_1|representation_2", "reasoning": "<brief sentence>"}}, "q8": {{"selection": "representation_1|representation_2", "reasoning": "<brief sentence>"}}, "q9": {{"selection": "representation_1|representation_2", "reasoning": "<brief sentence>"}}, "q10": {{"selection": "representation_1|representation_2", "reasoning": "<brief sentence>"}}, "q11": {{"selection": "representation_1|representation_2", "reasoning": "<brief sentence>"}}
}}
\end{lstlisting}
\end{tcolorbox}

\begin{table}[h]
    \centering
    \caption{Comparing user representations generated by our trained and untrained models along several dimensions. We report the percentage of times that GPT-5.4 mini selects the representation generated by \texttt{UserRepGen-4B} over the one generated by Qwen3.5-4B.}
    \label{tab:rep-analysis}
    \small
    \begin{tabular}{@{}p{0.6\linewidth}r@{}}
        \toprule
        \textbf{Question} & \textbf{\texttt{UserRepGen-4B} Selected (\%)} \\
        \midrule
        \multicolumn{2}{@{}l}{\textit{Task description}} \\
        \quad Captures constraints introduced in later turns
            & 63.6 \\
        \quad Captures shifts in direction or goal
            & 61.0 \\
        \quad Captures multiple objectives or subtasks
            & 58.3 \\
        \quad Has less repetition
            & 33.1 \\
        \midrule
        \multicolumn{2}{@{}l}{\textit{Behavior description}} \\
        \quad Is more specific and actionable
            & 87.7 \\
        \quad Uses more useful negative and contrastive instructions
            & 91.6 \\
        \quad Captures the user's response policy
            & 83.1 \\
        \quad Captures the user's communication style
            & 88.1 \\
        \quad Captures the user's engagement patterns
            & 90.0 \\
        \quad Captures how the user makes requests
            & 83.9 \\
        \quad Captures how the user provides context
            & 85.9 \\
        \bottomrule
    \end{tabular}
\end{table}

Our results are presented in Table~\ref{tab:rep-analysis}. GPT-5.4 mini selects the task description from \texttt{UserRepGen-4B} as better capturing constraints introduced in later turns in 63.6\% of comparisons, shifts in direction or goal in 61.0\%, and multiple objectives in 58.3\%. Together, these results support our qualitative observation that training produces more comprehensive task descriptions. However, \texttt{UserRepGen-4B} is selected as less repetitive in only 33.1\% of comparisons, indicating that Qwen3.5-4B usually produces less repetitive task descriptions. For behavior descriptions, GPT-5.4 mini selects the description from \texttt{UserRepGen-4B} as containing more useful negative and contrastive instructions in 91.6\% of comparisons, better capturing user engagement patterns in 90.0\%, better capturing communication style in 88.1\%, and being more specific and actionable in 87.7\%. It also selects \texttt{UserRepGen-4B} as better describing how the user provides context in 85.9\%, how the user makes requests in 83.9\%, and how the user responds to the assistant in 83.1\%. These results support our qualitative observation that training produces behavior descriptions that are more specific and actionable, and that capture dimensions of user behavior previously overlooked.

\subsection{User Representation Examples}

We provide two examples of user representations generated by Qwen3.5-4B and \texttt{UserRepGen-4B}. Yellow highlighting marks the differences most relevant to our analysis.

\paragraph{Example 1.} This example illustrates how \texttt{UserRepGen-4B} generates task descriptions that capture the full sequence of user requests, whereas the base model summarizes the conversation at a high-level. The behavior description replaces broad terms such as ``curious'', ``direct'', and ``inquisitive'' with observable patterns that a simulator can reproduce, such as specifying that the user asks short factual questions, provides no initial context, and pivots to new angles rather than giving feedback.

\begin{tcolorbox}[
  colback=gray!5,
  colframe=gray!75!black,
  title=User representation generated by Qwen3.5-4B,
  fonttitle=\bfseries,
  left=2mm,
  right=2mm,
  top=1mm,
  bottom=1mm,
  arc=2mm,
  boxrule=0.5pt,
  breakable
]
\small
\noindent\textit{Task description.}
The user seeks to understand the core ideas and historical impact of a specific philosopher, Čaadaev, and subsequently explores the
philosophical connections between him and other notable figures like Dostoevsky.

\smallskip
\noindent\textit{Behavior description.}
The user initiates inquiries with \repchange{broad, open-ended questions about a topic's fundamental nature and relevance}. They
progressively narrow the scope by requesting specific follow-up information on influences and related concepts. Their engagement is
characterized by \repchange{a high degree of curiosity} and a desire for concise summaries, often explicitly requesting structural
constraints (such as a sentence limit) when formulating their requests. The communication style is \repchange{direct and inquisitive},
treating the assistant as a knowledge repository without providing extensive initial context.
\end{tcolorbox}

\begin{tcolorbox}[
  colback=blue!3,
  colframe=blue!50!black,
  title=User representation generated by \texttt{UserRepGen-4B},
  fonttitle=\bfseries,
  left=2mm,
  right=2mm,
  top=1mm,
  bottom=1mm,
  arc=2mm,
  boxrule=0.5pt,
  breakable
]
\small
\noindent\textit{Task description.}
The user seeks to understand the philosophical contributions of Ivan Čaadaev and their influence on later thinkers, specifically focusing
on Fyodor Dostoevsky. The interaction involves \repchange{a series of sequential inquiries about Čaadaev's core themes, modern relevance,
intellectual lineage, and a specific request for a concise summary of Dostoevsky's philosophy}.

\smallskip
\noindent\textit{Behavior description.}
The user engages in a multi-turn dialogue characterized by \repchange{short, direct, and factual questions that progressively narrow down the scope of inquiry} from general biography to specific thematic analysis and finally to a constrained output format (two sentences). They
decompose complex topics into \repchange{discrete sub-questions without providing initial context or background information}, relying instead on the assistant's responses to inform subsequent queries. The communication style is \repchange{terse and impersonal, avoiding
conversational fillers, emotional expressions, or requests for formatting details like markdown headers}. \repchange{The user does not provide feedback on previous answers but immediately pivots to new angles or deeper dives into related figures based on the assistant's
output}. They maintain a consistent pattern of asking distinct follow-ups rather than combining multiple goals into single prompts or
requesting broad overviews before drilling down. The engagement level is moderate; they do not terminate the conversation early nor do they
expand significantly beyond the immediate topic area, preferring a linear progression of specific facts and summaries.
\end{tcolorbox}

\medskip
\paragraph{Example 2.} In this example, the behavior description from \texttt{UserRepGen-4B} specifies how the user formulates requests: short, direct questions, often phrased as yes/no inquiries, with little upfront context. It also describes how the user responds to the assistant, noting that they neither critique nor refine previous answers but instead use them to guide the next question. The base model, by contrast, relies on broad characterizations.

\begin{tcolorbox}[
  colback=gray!5,
  colframe=gray!75!black,
  title=Representation generated by Qwen3.5-4B,
  fonttitle=\bfseries,
  left=2mm,
  right=2mm,
  top=1mm,
  bottom=1mm,
  arc=2mm,
  boxrule=0.5pt,
  breakable
]
\small
\noindent\textit{Task description.}
The user is investigating the safety, ethical implications, and management challenges of nuclear energy, specifically focusing on waste
disposal methods. They are exploring alternatives to traditional nuclear power and considering how artificial intelligence can contribute
to solving complex environmental and safety problems.

\smallskip
\noindent\textit{Behavior description.}
The user engages in a \repchange{structured inquiry} by posing direct questions that progressively narrow down from general safety concerns
to specific technical solutions and ethical considerations. They actively seek alternative perspectives and practical applications (such as
AI's role) rather than accepting single answers. Their communication style is \repchange{concise and focused on high-level concepts},
demonstrating a degree of engagement that balances curiosity with a desire for actionable information.
\end{tcolorbox}

\begin{tcolorbox}[
  colback=blue!3,
  colframe=blue!50!black,
  title=Representation generated by \texttt{UserRepGen-4B},
  fonttitle=\bfseries,
  left=2mm,
  right=2mm,
  top=1mm,
  bottom=1mm,
  arc=2mm,
  boxrule=0.5pt,
  breakable
]
\small
\noindent\textit{Task description.}
The user is exploring the safety, morality, and future implications of nuclear energy, specifically focusing on waste disposal methods and
seeking potential AI-driven solutions for these challenges.

\smallskip
\noindent\textit{Behavior description.}
The user engages in a multi-turn dialogue with a distinct communication style characterized by \repchange{short, direct questions often
phrased as simple yes/no inquiries or brief statements}. They decompose complex topics into \repchange{sequential sub-questions without
providing extensive background context upfront}, relying instead on the assistant's responses to guide subsequent turns. The user maintains
\repchange{a low level of engagement per turn, avoiding elaboration, justification, or follow-up analysis beyond the immediate question}.
They do not provide feedback, corrections, or alternative perspectives; their interaction pattern is \repchange{strictly linear and
interrogative}. The user consistently \repchange{asks new, related questions rather than refining previous ones or requesting specific
formatting changes}.
\end{tcolorbox}

\section{Measuring the Gap Between Distributions of Original and Simulated User Behaviors}
\label{appendix:distributional-gap}

To evaluate how well user representations preserve behavioral diversity at the population level, we measure the distributional gap between original and simulated user behaviors. We follow the pipeline and hyperparameters of \citet{mehri2026measuring}, which we detail below.

\paragraph{User behavior descriptions.}
We use Qwen3.5-122B-A10B-FP8 to describe the user behavior in each original and simulated conversation along six behavioral dimensions: requests, responses, context, communication style, and dialog acts under the Dialog Act Markup in several Layers (DAMSL) framework and Schema-Guided Dataset (SGD) taxonomies. Each prompt instructs the model to describe general behavioral patterns while excluding task-specific details. We use the following prompts:

\begin{tcolorbox}[
    colback=gray!5,
    colframe=gray!75!black,
    title=User Behavior Description: Requests Dimension,
    fonttitle=\bfseries,
    left=2mm,
    right=2mm,
    top=1mm,
    bottom=1mm,
    arc=2mm,
    boxrule=0.5pt,
    breakable
]
\begin{lstlisting}[
    basicstyle=\scriptsize\ttfamily,
    breaklines=true,
]
You are an expert analyzing user behaviour in human-AI conversations. The user has a goal, and the assistant helps them achieve it. Your task is to describe the user's behavior according to the criteria below.

# User Goal
{user_goal}
# User Utterances
{conversation_history}

# Analysis Criteria
1. Specification and Articulation
   - How specified are the requests? Is the first request underspecified, and clarified in subsequent turns? Or are the requests exhaustive? 
   - What type of information is left underspecified/specified? (e.g. constraints, edge cases, context, output format, etc.)
   - Does the user explicitly command specific tasks, or do they rely on indirect cues (e.g., presenting an error without explicitly asking for a fix)?

2. User Goal Decomposition
   - How is the user goal decomposed across utterances?
      - Single-shot: The entire goal is expressed in one utterance with no further decomposition.
      - Top-down: High-level goal stated upfront, then progressively refined or elaborated.
      - Bottom-up: Individual preconditions or sub-tasks addressed first, building toward the overall goal.
      - Chained: Each request builds purely on the immediate previous turn rather than referring back to a central goal.

3. Relevance to Goal
   - Are the requests directly related to the user goal? Or does the user introduce secondary/perpendicular/emergent needs? 
   - What functions do the requests serve beyond achieving the user goal? (setting context, probing AI capabilities, verifying intermediate outputs, logistics, troubleshooting, side-effects, exploring related sub-tasks, etc.)

# Instructions
- Generate terse, concise bullet points, not full sentences.
- Abstract away from the specific topic/domain
- IMPORTANT:Do NOT use task-specific terms (e.g., "coding," "booking," "Python", "CSV"). Use generic substitutes (e.g., "executing a task," "providing constraints," "the target artifact").

Output a valid JSON object using the exact format below. Do not include any text outside the JSON.

{{
   "specification_and_articulation": {{
      "specification_level": "how specified are requests?",
      "underspecified_aspects": "what types of information are left underspecified/specified? (provide high-level descriptions without task-specific details)",
      "articulation_mode": "how does the user articulate their requests (explicit directives, indirect cues, mixed)?"
   }},
   "goal_decomposition_strategy": "describe how the user goal decomposed across utterances?",
   "relevance_to_goal": {{
      "goal_adherence": "Are the user's requests directly related to the user goal? If not, what other functions do they serve? (provide high-level descriptions without task-specific details)"
   }}
}}
\end{lstlisting}
\end{tcolorbox}

\begin{tcolorbox}[
    colback=gray!5,
    colframe=gray!75!black,
    title=User Behavior Description: Responses Dimension,
    fonttitle=\bfseries,
    left=2mm,
    right=2mm,
    top=1mm,
    bottom=1mm,
    arc=2mm,
    boxrule=0.5pt,
    breakable
]
\begin{lstlisting}[
    basicstyle=\scriptsize\ttfamily,
    breaklines=true,
]
You are an expert analyzing user behaviour in human-AI conversations. The user has a goal, and the assistant helps them achieve it. Your task is to describe the user's behavior according to the criteria below.

# User Goal
{user_goal}

# Conversation
{conversation_history}

# Analysis Criteria
1. Engagement and Evaluation
   - Does the user engage with the agent responses, or ignore/skip over them?
   - How does the user evaluate the agent's output? (e.g., explicit validation, implicit acceptance by continuing, partial acceptance with corrections, rejection)
   - Does the user provide specific or actionable feedback, or only surface-level acknowledgment?

2. Response Composition
   - What types of actions are present in the user's responses? Are they reactive (e.g., validation, acknowledgment, answering agent questions, corrections, feedback), proactive (e.g., follow-up requests, new constraints/preferences, suggestions, questions), or self-directed (e.g., thinking out loud, expressing uncertainty, narrating their process)?

3. Steering Mechanism
   - Does the user steer through direct follow-up requests, or through indirect means? (e.g., asking questions that implicitly request action, expressing dissatisfaction without stating what to change, providing hints or examples rather than directives)
   - Does the user introduce new preferences, constraints, or feedback as part of their responses, effectively reshaping the task incrementally?

# Instructions
- Generate terse, concise bullet points, not full sentences.
- Abstract away from the specific topic/domain.
- IMPORTANT: Do NOT use task-specific terms (e.g., "coding," "booking," "Python", "CSV"). Use generic substitutes (e.g., "executing a task," "providing constraints," "the target artifact").

Output a valid JSON object using the exact format below. Do not include any text outside the JSON.

{{
   "engagement_and_evaluation": {{
      "engagement_level": "does the user engage with the agent's responses or skip over them?",
      "evaluation_mode": "how does the user evaluate the agent's output? (explicit validation, implicit acceptance, partial acceptance, rejection, etc.)",
      "feedback_specificity": "does the user provide specific/actionable feedback or only surface-level acknowledgment?"
   }},
   "response_composition": {{
      "action_types": "What types of actions are present in the user's responses? Are they reactive (e.g., validation, acknowledgment, answering agent questions, corrections, feedback), proactive (e.g., follow-up requests, new constraints/preferences, suggestions, questions), or self-directed (e.g., thinking out loud, expressing uncertainty, narrating their process)?"
   }},
   "steering_mechanism": {{
      "directness": "does the user steer through explicit follow-up requests or indirect means (questions, hints, expressed dissatisfaction)?",
      "incremental_reshaping": "does the user introduce new preferences/constraints/feedback that reshape the task within their responses?"
   }}
}}
\end{lstlisting}
\end{tcolorbox}

\begin{tcolorbox}[
    colback=gray!5,
    colframe=gray!75!black,
    title=User Behavior Description: Context Dimension,
    fonttitle=\bfseries,
    left=2mm,
    right=2mm,
    top=1mm,
    bottom=1mm,
    arc=2mm,
    boxrule=0.5pt,
    breakable
]
\begin{lstlisting}[
    basicstyle=\scriptsize\ttfamily,
    breaklines=true,
]
You are an expert analyzing user behaviour in human-AI conversations. The user has a goal, and the assistant helps them achieve it. Your task is to describe the user's behavior according to the criteria below.

# User Goal
{user_goal}

# Conversation
{conversation_history}

# Analysis Criteria
1. Context Richness
   - How much context does the user provide overall? Do they share background about themselves, their situation, what they've already tried, what they're struggling with, or what they're thinking?
   - Does the user provide all relevant context needed for the agent to help effectively, or is essential context left out?
   - Does the user provide context at all, or do they simply issue directives without situating the task?

2. Context Type
   - What types of context does the user provide? (e.g., personal background, goals/motivations, prior attempts, existing solutions, constraints, preferences, domain knowledge, emotional state, thought process)

3. Context Delivery
   - Does the user front-load all relevant context in their first message, or reveal it incrementally across turns? What types of context are introduced later? (e.g., constraints they forgot, preferences they didn't think to mention, background that becomes relevant as the task evolves)
   - Is incremental context revealed in response to agent questions, or volunteered unprompted?

# Instructions
- Generate terse, concise bullet points, not full sentences.
- Abstract away from the specific topic/domain.
- IMPORTANT: Do NOT use task-specific terms (e.g., "coding," "booking," "Python", "CSV"). Use generic substitutes (e.g., "executing a task," "providing constraints," "the target artifact").

Output a valid JSON object using the exact format below. Do not include any text outside the JSON.

{{
   "context_richness": {{
      "depth": "how much context does the user provide? do they share background, prior attempts, thought process, etc.?",
      "completeness": "does the user provide all relevant context or leave essential information out?",
      "contextualization_vs_directives": "does the user situate the task with context, or simply issue directives?"
   }},
   "context_type": "what types of context does the user provide? (personal background, goals/motivations, prior attempts, existing solutions, constraints, preferences, domain knowledge, emotional state, thought process, etc.)",
   "context_delivery": {{
      "distribution": "does the user front-load context or reveal it incrementally? What types of context are introduced in later turns?",
      "trigger": "is incremental context revealed in response to agent questions or volunteered unprompted?"
   }}
}}
\end{lstlisting}
\end{tcolorbox}

\begin{tcolorbox}[
    colback=gray!5,
    colframe=gray!75!black,
    title=User Behavior Description: Communication Style Dimension,
    fonttitle=\bfseries,
    left=2mm,
    right=2mm,
    top=1mm,
    bottom=1mm,
    arc=2mm,
    boxrule=0.5pt,
    breakable
]
\begin{lstlisting}[
    basicstyle=\scriptsize\ttfamily,
    breaklines=true,
]
You are an expert analyzing user behaviour in human-AI conversations. The user has a goal, and the assistant helps them achieve it. Your task is to describe the user's behavior according to the criteria below.

# User Goal
{user_goal}

# Conversation
{conversation_history}

# Analysis Criteria
1. Register and Tone
   - What is the user's register? (e.g., formal, casual, conversational, terse, professional)
   - What is the user's emotional tone? (e.g., neutral, frustrated, enthusiastic, impatient, apologetic, deferential)
   - Does the register/tone shift across the conversation? (e.g., starts polite but becomes curt after errors)

2. Verbosity and Structure
   - How verbose are the user's messages? Are they minimal and compressed, or expansive and elaborated?
   - Does the user use formatting conventions? (e.g., bullet points, numbered lists, code blocks, markdown, all caps, punctuation patterns)
   - How does message length and structure change across turns?

3. Social Conventions
   - Does the user use greetings, pleasantries, expressions of gratitude, or sign-offs?
   - Does the user treat the agent as a tool (purely transactional) or as a social interlocutor (rapport-building, politeness, humor)?

4. Request Framing
   - How does the user syntactically frame their requests? (e.g., imperative commands, questions, hedged suggestions, statements of problems without explicit asks)

# Instructions
- Generate terse, concise bullet points, not full sentences.
- Abstract away from the specific topic/domain.
- IMPORTANT: Do NOT use task-specific terms (e.g., "coding," "booking," "Python", "CSV"). Use generic substitutes (e.g., "executing a task," "providing constraints," "the target artifact").

Output a valid JSON object using the exact format below. Do not include any text outside the JSON.

{{
   "register_and_tone": {{
      "register": "what is the user's register? (formal, casual, conversational, terse, professional, etc.)",
      "emotional_tone": "what is the user's emotional tone? (neutral, frustrated, enthusiastic, impatient, apologetic, deferential, etc.)",
      "tone_shifts": "does the tone shift across the conversation, and in response to what?"
   }},
   "verbosity_and_structure": {{
      "verbosity": "how verbose are the user's messages? minimal and compressed, or expansive and elaborated?",
      "formatting": "does the user use formatting conventions? (bullet points, numbered lists, code blocks, markdown, etc.)",
      "evolution": "how does message length and structure change across turns?"
   }},
   "social_conventions": {{
      "politeness_markers": "does the user use greetings, pleasantries, gratitude, or sign-offs?",
      "agent_relationship": "does the user treat the agent as a tool (transactional) or as a social interlocutor (rapport-building, politeness, humor)?"
   }},
   "request_framing": "how does the user syntactically frame their requests? (imperative commands, questions, hedged suggestions, problem statements without explicit asks, etc.)"
}}
\end{lstlisting}
\end{tcolorbox}

\begin{tcolorbox}[
    colback=gray!5,
    colframe=gray!75!black,
    title=User Behavior Description: DAMSL Dialog Acts Dimension,
    fonttitle=\bfseries,
    left=2mm,
    right=2mm,
    top=1mm,
    bottom=1mm,
    arc=2mm,
    boxrule=0.5pt,
    breakable
]
\begin{lstlisting}[
    basicstyle=\scriptsize\ttfamily,
    breaklines=true,
]
You are an expert analyzing user behaviour in human-AI conversations. Your task is to analyze a specific user utterance in a multi-turn conversation and provide a description of the user's behaviour.

# Conversation History
{conversation_history}

# Target Utterance
Read the conversation history for context, but your analysis MUST strictly apply ONLY to this utterance:
{target_user_utterance}


# Dialogue Act Markup in Several Layers (DAMSL) Annotation Framework
DAMSL defines utterances based on the intentions of the speaker. Each utterance is analyzed along several dimensions.
1. INFORMATION LEVEL: Characterizes the semantic content of the utterance:
    - **Task**: Directly advances (or attempts to advance) the goals of the domain task.
    - **Task-management**: Addresses the problem-solving process/procedure rather than performing the task itself.
    - **Communication-management**: Concerned exclusively with maintaining the communication channel.
2. FORWARD LOOKING FUNCTION: Characterizes what effect the utterance has on the subsequent dialog:
    - **Statement**: Makes a claim (Assert, Re-assert, Other-statement).
    - **Info-request**: Asks for information.
    - **Influencing-addressee-future-action**: Directly influences the addressee's future non-communicative actions, directs or suggests the addressee to perform an action.
    - **Committing-speaker-future-action**: Potentially commits the speaker to some future action.
    - **Conventional**: Conventional conversational actions such as greeting, farewells, thanking, or responding to thanks.
3. BACKWARD LOOKING FUNCTION: Characterizes how the current utterance relates to the previous discourse:
    - **Agreement**: Accept, reject, partial accept/reject, holds off a response.
    - **Answer**: Provides requested information.
    - **Understanding**: Signals comprehension (acknowledgement, repetition/reformulation, collaborative completion) or lack thereof.


# Output Instructions
Analyze the target utterance along DAMSL's dimensions and write a concise, 1 sentence description of the user's behaviour and its functional role in the target utterance:
- Generate terse, concise bullet points, not full sentences.
- Abstract away from the specific topic/domain
- IMPORTANT:Do NOT use task-specific terms (e.g., "coding," "booking," "Python", "CSV"). Use generic substitutes (e.g., "executing a task," "providing constraints," "the target artifact").

Output only the concise description as plain text, without any other text or formatting.
\end{lstlisting}
\end{tcolorbox}

\begin{tcolorbox}[
    colback=gray!5,
    colframe=gray!75!black,
    title=User Behavior Description: SGD Dialog Acts Dimension,
    fonttitle=\bfseries,
    left=2mm,
    right=2mm,
    top=1mm,
    bottom=1mm,
    arc=2mm,
    boxrule=0.5pt,
    breakable
]
\begin{lstlisting}[
    basicstyle=\scriptsize\ttfamily,
    breaklines=true,
]
You are an expert analyzing user behaviour in human-AI conversations. Your task is to analyze a specific user utterance in a multi-turn conversation and classify the user's dialogue act(s).

# Conversation History
{conversation_history}

# Target Utterance
Read the conversation history for context, but your analysis MUST strictly apply ONLY to this utterance:
{target_user_utterance}


# Dialogue Act Taxonomy (adapted from Schema-Guided Dialogue)
Classify the target utterance using one or more of the following user dialogue acts:

1. INFORM: User provides information, states preferences, or supplies constraints to the agent.
2. REQUEST: User asks the agent for specific information or details.
3. AFFIRM: User agrees with, confirms, or accepts a proposition, value, or suggestion made by the agent.
4. NEGATE: User disagrees with, denies, or rejects a proposition, value, or suggestion made by the agent.
5. SELECT: User chooses between multiple options or alternatives presented by the agent.
6. INFORM_INTENT: User states, introduces, or shifts to a new goal or task objective.
7. AFFIRM_INTENT: User confirms or agrees with a goal or task objective suggested by the agent.
8. NEGATE_INTENT: User rejects or denies a goal or task objective suggested by the agent.
9. REQUEST_ALTS: User asks for different or additional options beyond what has been presented.
10. THANK: User expresses gratitude or appreciation.
11. GOODBYE: User signals the end of the conversation or a conversational closing.
12. GREET: User initiates the conversation with a greeting or opening.


# Output Instructions
- Assign ALL applicable dialogue acts to the target utterance (utterances may have multiple acts).

Output the dialogue acts, separated by commas, without any other text or formatting.
\end{lstlisting}
\end{tcolorbox}

\paragraph{Embedding and quantization.} For each conversation, we concatenate the descriptions across all six behavioral dimensions and embed the resulting text using Qwen3-Embedding-8B. We retain the first 1,024 embedding dimensions and apply $\ell_2$ normalization. For each comparison between original and simulated conversations, we pool the two sets of embeddings and apply PCA to retain 90\% of the variance. We then quantize the embeddings using $k$-means with $k=500$ clusters, using five restarts with up to 500 iterations each and retaining the solution with the lowest clustering objective. The cluster counts are used to create discrete distributions for the original and simulated user behaviors.

\end{document}